\RequirePackage{fix-cm}
\documentclass[twocolumn]{svjour3}          
\smartqed  
\usepackage{graphicx}
\usepackage[numbers]{natbib}
\usepackage{multirow}
\usepackage{amsmath}
\usepackage{amssymb}
\usepackage{bbm}
\usepackage{bm}
\usepackage{booktabs}
\usepackage{array} 
\usepackage{blindtext}
\usepackage{comment}

\usepackage{bm}
\usepackage{listings}
\usepackage{subcaption}
\usepackage[table,svgnames, dvipsnames]{xcolor}
\usepackage{mathtools}

\definecolor{myred}{rgb}{1,0,0}
\definecolor{mygreen}{rgb}{0,0.6,0}
\definecolor{mygray}{rgb}{0.5,0.5,0.5}
\definecolor{mymauve}{rgb}{0.58,0,0.82}

\newcommand{\gr}[1]{\small{\textcolor{ForestGreen}{(\textbf{#1})}}}

\newcommand{\bluegr}[1]{\small{\textcolor{RoyalBlue}{(\textbf{#1})}}}

\newcommand{\cmark}{\ding{51}}%
\newcommand{\xmark}{\ding{55}}%

\usepackage{tabularx, booktabs, makecell}
\newcolumntype{Y}{>{\centering\arraybackslash}m{2.2cm}}
\newcolumntype{C}{>{\centering\arraybackslash}X}

\usepackage{xspace}

\usepackage[utf8]{inputenc} 
\usepackage[T1]{fontenc}    
\usepackage{booktabs}       
\usepackage{amsfonts}       
\usepackage{nicefrac}       
\usepackage{microtype}      

\usepackage{graphicx}
\usepackage{amsmath,amssymb} 
\usepackage{caption} 
\usepackage{hyperref}
\usepackage{wrapfig}
\usepackage[american]{babel}
\usepackage{csquotes}

\usepackage{bm}
\usepackage{paralist,algorithm}
\usepackage{multirow}

\usepackage[noend]{algpseudocode}

\usepackage{accents}

\definecolor{Gray}{gray}{0.85}
\definecolor{LightCyan}{rgb}{0.88,1,1}

\makeatletter
\DeclareRobustCommand\onedot{\futurelet\@let@token\@onedot}
\def\@onedot{\ifx\@let@token.\else.\null\fi\xspace}

\makeatother

\usepackage{pifont}

\usepackage[misc]{ifsym}
\usepackage{color, colortbl}
\definecolor{citecolor}{HTML}{0071bc}
\definecolor{tabhighlight}{HTML}{e5e5e5}

\makeatletter
\renewcommand\paragraph{
  \@startsection{paragraph} 
  {4} 
  {\z@} 
  {.5em \@plus1ex \@minus.2ex} 
  {-.5em} 
  {\normalfont\normalsize\bfseries} 
}
\makeatother

\begin{document}
\sloppy

\title{LW2G: Learning Whether to Grow for Prompt-based Continual Learning
}

\author{Qian Feng \and 
        Da-Wei Zhou  \and
        Hanbin Zhao \and
        Chao Zhang \and
        Jiahua Dong \and
        Dengxin Dai \and
        Hui Qian 
}

\institute{Qian Feng \at
            College of Computer Science and Technology, Zhejiang University, China \\
              \email{fqzju@zju.edu.cn}
           \and
           Da-Wei Zhou \at
             State Key Laboratory for Novel Software Technology, Nanjing University, China \\
              \email{zhoudw@lamda.nju.edu.cn}
           \and
           Hanbin Zhao \at
            College of Computer Science and Technology, Zhejiang University, China \\
              \email{zhaohanbin@zju.edu.cn}
           \and
           Chao Zhang \at
            College of Computer Science and Technology, Zhejiang University, China \\
              \email{zczju@zju.edu.cn}
           \and
           Jiahua Dong \at
            Mohamed bin Zayed University of Artificial Intelligence, Abu Dhabi \\
              \email{dongjiahua1995@gmail.com}
           \and
           Dengxin Dai \at
            Huawei Zurich Research Center, Switzerland \\
              \email{dengxin.dai@huawei.com}
           \and
           Hui Qian \at
            College of Computer Science and Technology, Zhejiang University, China \\
              \email{qianhui@zju.edu.cn}
           \and 
           Hanbin Zhao is corresponding author.
           }

\date{Received: date / Accepted: date}

\maketitle

\begin{abstract}
  Recent Prompt-based Continual learning (PCL) has achieved remarkable performance with pre-trained models. These approaches expand a prompt pool by adding a new set of prompts while learning and select the correct set during inference. Previous studies have revealed that learning task-wised prompt sets individually and low selection accuracy pose challenges to the performance of PCL. In this paper, we propose a plug-in method, \textbf{L}earning \textbf{W}hether \textbf{t}o \textbf{G}row \textbf{(LW2G)}, which leverages the disparities between tasks to form an effective and efficient prompt sets pool, thereby achieving intra-task knowledge sharing and cooperation and avoiding the unbounded increase in the cost of the prompt pool. Specifically, a shared set is utilized when several tasks share certain commonalities, and a new set is added when there are significant differences between the new and previous tasks. To achieve this, we develop a metric called Hinder Forward Capability (HFC) to measure the hindrance imposed on learning new tasks by surgically modifying the original gradient onto the orthogonal complement of the old feature space. With HFC, an automated scheme, Dynamic Growing Approach, adaptively learns whether to grow with a dynamic threshold. Furthermore, we design a gradient-based constraint to ensure consistency between the updating prompts and pre-trained knowledge. Extensive experiments show the effectiveness of our method. Code is available at \url{https://github.com/RAIAN08/LW2G}.
\end{abstract}


\section{Introduction}
\label{sec:introduction}

\begin{figure*}[t]
    \centering
    \vspace{-2mm}
    \includegraphics[width=1\textwidth]{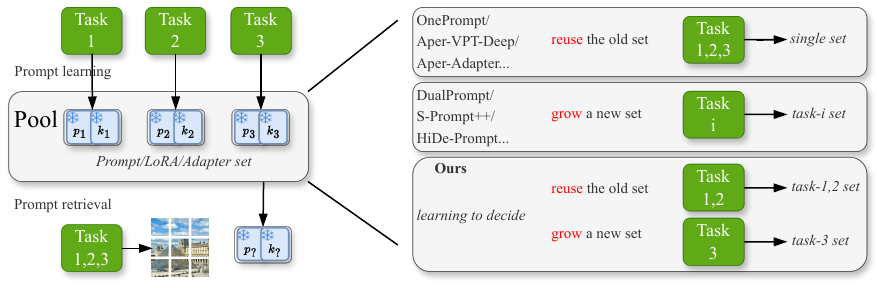}
    \vspace{-6mm}
    \caption{Compared with the latest PCL methods, our method addresses a timely and unresolved question: \textbf{when there are three tasks, what is the minimum number of prompt sets required?} Unlike Aper-VPT-Deep, which adopts a coarse strategy of using a single prompt set and suffers from limited representation capacity, and unlike DualPrompt, which incurs linear growth of the prompt pool and adopts separate learning that hinders inter-task knowledge sharing and collaboration, our method dynamically determines whether to grow a new prompt set or reuse an existing one based on task disparities. This results in an effective and efficient prompt set pool.}
  \label{first_pic}
   \vspace{-4mm}
\end{figure*}

Compared  to learning in stationary scenarios, Continual Learning (CL) equips systems with the ability to learn in non-stationary environments, which is a core step toward achieving human-level intelligence and human-like adaptation. This paradigm has been widely adopted in computer vision (e.g., object detection \cite{feng2022overcoming,zhao2022static}, semantic segmentation \cite{yan2021framework,zhu2023continual,zhao2022rbc}), natural language processing (e.g., relation extraction \cite{hu2022improving}, neural machine translation \cite{zhang2022continual}), and multi-modal scenarios (e.g., cross-modal retrieval \cite{zhao2021video}, visual question answering \cite{lei2023symbolic,qian2023decouple}). In this paradigm, Deep Neural Networks (DNNs) need to learn from sequential tasks while retaining past knowledge and acquiring novel knowledge. However, simply utilizing standard optimization methods \cite{diederik2014adam, ruder2016overview} for training DNNs inevitably erases the parametric representations of old tasks with new input representations during updating. Therefore, a well-known issue \textit{catastrophic forgetting} (CF) arises \cite{french1999catastrophic, ramasesh2021effect, mccloskey1989catastrophic, rebuffi2017icarl, lewandowsky1995catastrophic}, where DNNs suffer severe performance degradation on old tasks due to the absence of old data and domain shift in data distributions, making CL an extremely challenging problem.

Recently, \textbf{Prompt-based Continual Learning (PCL)} offers fresh insights into addressing catastrophic forgetting (CF) \cite{wang2024hierarchical, douillard2022dytox, smith2023coda, zhou2025revisiting, liu2024compositional, wang2022s, wang2022dualprompt, zhou2022learning}. These methods leverage frozen Pre-Trained Models (PTMs) instead of training from scratch and adopt Parameter-Efficient Fine-Tuning (PEFT) techniques \cite{zhu2023prompt, dettmers2024qLoRA, wang2020k, houlsby2019parameter, jia2022visual, hu2021LoRA}, such as prompt, LoRA. Specifically, PCL typically consists of two stages: (a) \textit{prompt learning}, which learns a task-specific set of prompts to conditionally guide the PTM for the current task—these prompts are stored in an expanding prompt pool; and (b) \textit{prompt retrieval}, which predicts the task to which each test sample belongs and selects the corresponding prompt set. Recent studies \cite{wang2024hierarchical, huang2024ovor, tran2023koppa} have shown that \textbf{the accuracy of prompt selection significantly affects overall performance}, as assigning an incorrect prompt set to a test sample can lead to a notable performance decline. Furthermore, learning each task with an \textbf{independent prompt set or LoRA set neglects the potential for inter-task knowledge sharing and collaboration}, thereby limiting the representational capacity and effectiveness \cite{yu2024boosting, rypesc2024divide}. 



Prior works address the two issues in different ways, as illustrated in Figure \ref{first_pic}. For example, methods such as OnePrompt and Aper-VPT-Deep \cite{zhou2025revisiting} abandon the prompt pool and instead adopt a single prompt/LoRA set throughout the entire continual learning process. They enforce this single set to be effective for all tasks by adding regularization terms. However, a natural drawback is that as the number of tasks increases or the task diversity grows, a single set inevitably suffers from limited representation capacity. Another line of methods, represented by S-Prompt \cite{wang2022s} and HiDe-Prompt \cite{wang2024hierarchical}, focuses on improving the prompt selection strategy, such as using KNN or an auxiliary classification head. However, as the prompt pool expands, they still suffer from low selection accuracy, which leads to poor performance.

One simple solution to these problems is to mimic humans' integration of information \cite{roediger1995creating, hunt2006concept, arndt2006distinctive}. For instance, when several tasks share certain commonalities, they can utilize a shared set of prompts. In cases where tasks differ substantially, it is more appropriate to introduce a new set. Thus, by adaptively learning whether to grow a new set for PCL, the amount of selectable options is reduced, and the divergence between prompt sets is increased, thereby improving selecting accuracy. Furthermore, aggregating multiple tasks' knowledge into a single and shared set can also facilitate its representational capacity. Nevertheless, establishing suitable metrics to measure this commonality and obtaining task information \emph{a prior} -- all of which are challenging in practice. Moreover, gradually integrating knowledge from multiple tasks into a single and shared prompt set also presents an unresolved question, as knowledge from different tasks may interfere with each other during sequential learning, especially when the data of previous tasks is no longer accessible, potentially leading to forgetting or representation collapse.

\begin{figure}[t]
  \centering
   \includegraphics[width=1\linewidth]{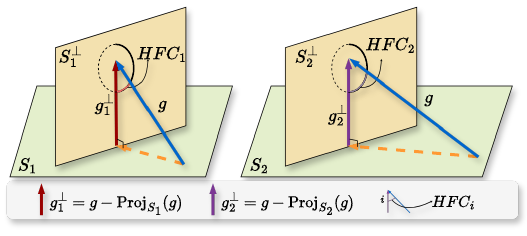}
   \caption{Illustration of HFC. $\mathcal{S}_{i},i=1,2$ represents the feature space spanned by the old task $i$, while $\mathcal{S}_{i}^{\perp}$ denotes the orthogonal complement of $\mathcal{S}_{i}$. $\bm{g}$ is the original gradient on learning task $3$. Then, $\text{HFC}(\bm{g}, \bm{g}_{i}^{\perp})$ is denoted as $\text{HFC}_{i}$.}
   \label{gpcl_pic}
\end{figure}

Thanks to Gradient Projection-based Continual Learning (GPCL) \cite{zhao2023rethinking, saha2021gradient, lopez2017gradient}, which proposes that learning would not forget if the updated gradient is orthogonal to the feature space spanned by old tasks (denoted as the \emph{orthogonal condition}), we adopt the \emph{orthogonal condition} in GPCL to integrate the knowledge from multiple tasks into a single set of prompts. Specifically, in Figure \ref{gpcl_pic}, the gradient $\bm{g}$ of task $3$ is modified to its projection $\bm{g}^{\perp}_{1}$(or $\bm{g}^{\perp}_{2}$) onto $\mathcal{S}_{1}^{\perp}$(or $\mathcal{S}_{2}^{\perp}$), and $\bm{g}^{\perp}_{1}$(or $\bm{g}^{\perp}_{2}$) serves as the real gradient for updating parameters, thereby posing no forgetting of old knowledge in task $1$ (or old knowledge of task $2$). Furthermore, inspired by \cite{zhao2023rethinking}, which reveals the hindrance encountered when learning a new task under \textit{the orthogonal condition}, we first theoretically analyze the magnitude of this hindrance and introduce a novel metric to evaluate it, called \textbf{Hinder Forward Capability (HFC)}. We then utilize the proposed HFC metric to address the dilemma of whether \emph{to grow} (i.e., initializing a new set of prompts) or \emph{not to grow} (i.e., reusing an old set of prompts from the pool). Specifically, \emph{HFC is calculated as the angle $\theta$ between the gradient of the new task, $\bm{g}$, and its projection onto the orthogonal complement of the feature space of the old task $i$, $\bm{g}^{\perp}_{i}$}, thereby measuring the degree of difference between the modified gradient and the original gradient, that is, the extent to which the modified gradient hinders the learning of the new task. As illustrated in Figure \ref{gpcl_pic}, $\text{HFC}_{1}<\text{HFC}_{2}$ implies that the hindrance to learning task $3$ on the set of prompts for task $2$ (i.e., $\bm{p_{2}}$) is larger than that on the set of prompts for task $1$ (i.e., $\bm{p_{1}}$) when learning task $3$ under the \emph{orthogonal condition}. In summary, when the hindrance to learning task $3$ on all the existing prompt set ($\bm{p_{1}}$ and $\bm{p_{2}}$) is severe, PCL should choose \emph{to grow} a new set; conversely, it tends \emph{not to grow} and chooses to reuse an old prompt set that learning task $3$ under the \emph{orthogonal condition} faces the least hindrance.

Based on the analysis, we propose a plug-in method within PCL to \textbf{L}earn \textbf{W}hether \textbf{t}o \textbf{G}row \textbf{(LW2G)}, consisting of two components: {Dynamic Growing Approach (DGA)} and {Consistency with Pre-trained Knowledge (CPK)}. DGA is an automated scheme to learn whether \emph{to grow} (adopt a new set of prompts and store it in the pool) or \emph{not to grow} (utilize an existing set of prompts from the pool) for new tasks based on the introduced HFC metric. Specifically, to incorporate knowledge from multiple tasks into a single set of prompts, we first employ the \emph{orthogonal condition} to learn new tasks without forgetting and calculate the hindrance to learning with each set in the prompt pool through HFC. Meanwhile, we consider an ideal scenario to generate a dynamic threshold, which learn the new task on the pre-trained knowledge feature space $\mathcal{S}^{\text{pre}}$ without any obstacles from old tasks. DGA chooses \emph{to grow} if all HFC values are above this threshold, indicating that learning with each set in the pool encounters excessive hindrance. Conversely, DGA chooses \emph{not to grow} by selecting the old set of prompts with the minimum HFC and learning the new task under the \emph{orthogonal condition}. CPK aims to balance the disruption to pre-trained knowledge caused by continual learning on new tasks and the reduced plasticity brought by strict orthogonality to the entire pre-trained feature space $\mathcal{S}^{\text{pre}}$. Therefore, we propose applying a soft constraint to the gradient when learning new tasks, aiming to align the gradient direction as closely as possible with the feature space of the pre-trained knowledge, ensuring consistency between prompt updates and pre-trained knowledge. Our main contributions are threefold:
\setlength{\leftmargini}{6pt} 

\begin{itemize}
\item We are the first to introduce a method for controlling the growth of the prompt pool in existing PCL by a novel metric, HFC, which measures the hindrance to learning new tasks under the \textit{orthogonal condition}.
\item Based on HFC, we develop an automated learning scheme, LW2G, which learns whether to grow or not to grow. We aim to form an effective and efficient prompt sets pool where each single set contains knowledge from multiple tasks, thereby serving as a shared prompt set to enhance efficacy.
\item LW2G is a plug-in method for existing PCL. We demonstrate its superiority across multiple benchmarks through extensive experiments on 12 baselines and 2 other plug-in methods.
\end{itemize}

\section{Related Work}
\vspace{-2mm}
\textbf{Continual Learning and Gradient Projection}
Numerous efforts have been made to alleviate the core issue of CF \cite{french1999catastrophic, ramasesh2021effect, mccloskey1989catastrophic}, which can be roughly categorized into three main categories: ($1$) Architecture-based \cite{rusu2016progressive,yoon2017lifelong,li2019learn,loo2020generalized,mallya2018packnet,serra2018overcoming,ke2020continual}, ($2$) Rehearsal-based \cite{buzzega2020dark,cha2021co2l,rebuffi2017icarl,wu2019large,ebrahimi2020adversarial,pham2021dualnet,zhao2021video,de2021continual,wang2018progressive}, and ($3$) Regularization-based \cite{kirkpatrick2017overcoming,zenke2017continual,sun2025continual,kong2023trust}. Among Regularization-based methods, GPCL methods \cite{zhao2023rethinking, saha2021gradient, lopez2017gradient, qiao2023prompt, lin2022trgp, lin2022towards, zhu2023prompt, yu2020gradient, wang2021training, duncker2020organizing, wang2023orthogonal, smith2023continual, chen2020just, chen2022class} focus on the gradient of the parameter. These methods project the gradient orthogonally to the feature space spanned by the old tasks, thereby not affecting the old knowledge. However, existing methods predominantly focus on non-expansion CL methods and CNN-based models, often applied in task-incremental learning scenarios. PGP \cite{qiao2023prompt} extended these approaches to PCL but merely combined them without deeper integration. In contrast, we delve deeper and are the first to explore the relationship and combination of GPCL and expansion CL methods (i.e., PCL) in terms of the degree of learning hindrance caused by strict \textit{orthogonality conditions} (via the proposed HFC metric).

\textbf{Prompt-based CL Methods and Transfer Learning.}
PCL garnered significant attention due to their utilization of PEFT techniques \cite{zhu2023prompt, dettmers2024qLoRA, wang2020k, houlsby2019parameter, jia2022visual, hu2021LoRA, yang2024rcs} to leverage PTMs, achieving rehearsal-free and promising performance \cite{wang2024hierarchical, douillard2022dytox, smith2023coda, zhou2025revisiting, liu2024compositional, wang2022s, wang2022dualprompt, zhou2022learning, qiao2023prompt, Wang_2022_CVPR, huang2024ovor, zhou2024expandable, zhou2024continual, zhou2023learning,kim2025one, gao2024consistent, smith2023continual, liang2024infLoRA}. Among them, Aper-VPT-Deep \cite{zhou2025revisiting} and OnePrompt \cite{huang2024ovor} use only a single prompt set with regularization terms avoiding the \textit{prompt retrieval} stage, DualPrompt \cite{wang2022dualprompt} proposed partitioning the knowledge of tasks into general and specific parts, and learns them with $g$-prompt and $e$-prompt, respectively. Similarly, C-Prompt \cite{liu2024compositional}, S-liPrompt and S-iPrompt \cite{wang2022s} addressed Domain-CL by leveraging Vision-Language Models (VLMs) to further enhance the learning ability. CODA-Prompt \cite{smith2023coda}, S-Prompt++ \cite{wang2024hierarchical} and HiDe-Prompt \cite{wang2024hierarchical} improved \emph{prompt retrieval} stage through \emph{attention mechanisms}, KNN and auxiliary adapter classifiers, respectively. Additionally, recent studies show that fine-tuning downstream tasks or conducting continual learning with PTMs—even when the PTM is frozen—often leads to overfitting due to the relatively limited amount of training data, thereby degrading the pre-trained knowledge \cite{lee2023read, li2024coleclip, zheng2023preventing, zhu2023prompt}.

\vspace{-3mm}
\section{Preliminaries and Notations}

\paragraph{Continual Learning}
Assume there is a sequence of tasks and their corresponding training datasets $\left\{\mathcal{D}^{i}, i=1, 2,\ldots \right\}$ without overlapping classes, where $\mathcal{D}^{t} = \left\{\left(\bm{x}_{i,t},\bm{y}_{i,t}\right)\right\}_{i=1}^{n_{t}}$ belongs to the task $t$. We denote the DNN as $\mathcal{W}=\left\{\theta^{l}\right\}_{l=1}^{L}$, where $\theta^{l}$ is the weight of layer $l$. Given a training sample $\bm{x}_{i,t}$, we denote $\bm{x}_{i,t}^{l}$ as the input of layer $l$ and the output is $\bm{x}_{i,t}^{l+1}=f^{l}\left({\theta}^{l}, \bm{x}_{i,t}^{l}\right)$, where $f^{l}$ is the operation of layer $l$. We simplify the loss function for learning task $t$ as $\mathcal{L}_{t}(\mathcal{D}^{t})$ and $\mathcal{W}_{t}=\left\{\theta^{l}_{t}\right\}_{l=1}^{L}$ as the DNN after training on task $t$.

\paragraph{Gradient Projection Continual Learning}
\label{pre_gpcl_methods}
First, for $\bm{A} \in \mathbb{R}^{m \times n}$ and a subspace $\mathcal{S}$ in Euclidean space with its orthogonal bases $\bm{B} \in \mathbb{R}^{n \times d}$, the projection of $\bm{A}$ onto the subspace $\mathcal{S}$ is denoted as follows:
\begin{align}
    \text{Proj}_{\mathcal{S}}\left(\bm{A}\right) = \bm{A}\bm{B}\left(\bm{B}\right)^{T},
\end{align}
where $\left( \cdot \right)^{T}$ is the matrix transpose. Then, following \cite{saha2021gradient}, we briefly introduce how GPCL reduces the interference of old knowledge when learning new tasks. Specifically, the total process involves two stages.

\textbf{Stage ($1$) Building of the new feature space.}
After leaning task $1$, GPCL first constructs a representation matrix for layer $l$ as $\bm{R}_{1}^{l} \in \mathbb{R}^{N \times d}$ from task $1$ only ($d$ is the output dimension of layer $l$). Next, Singular Value Decomposition (SVD) is performed on $\bm{R}_{1}^{l}$ followed by its $k$-rank approximation $\left(\bm{R}_{1}^{l}\right)_{k}$ with threshold, $\epsilon$:
    \begin{align}
        ||(\bm{R}_{1}^{l})_{k}||_{F}^{2} \geq \epsilon ||\bm{R}_{1}^{l}||_{F}^{2}.
    \end{align}
Therefore, the feature space for layer $l$ spanned by task $1$ is built by $\mathcal{S}_{1}^{l}=\text{span}\left\{\bm{B}_{1}^{l}\right\}$, where $\bm{B}_{1}^{l}$ is the orthogonal bases for $\mathcal{S}_{1}^{l}$. And $\mathcal{S}_{1}^{l}$ is stored in memory $\mathcal{M}=\left\{\mathcal{S}_{1}^{l}\right\}$.

When learning task $2$, the gradient of layer $l$ is denoted as $\bm{g} = \nabla_{\theta^{l}}\mathcal{L}_{2}$. As illustrated in Figure \ref{gpcl_pic}, GPCL modify the gradient as follows:
\begin{align}
    \bm{g}^{\perp}_{1} = \text{Proj}_{\mathcal{S}_{1}^{\perp}}(\bm{g}),
\end{align}
where $\mathcal{S}_{1}^{\perp}$ is the orthogonal complement of $\mathcal{S}_{1}^{l}$ and $\bm{g}^{\perp}_{1}$ serves as the real gradient for updating layer $l$. Let $\Delta \theta_{1}^{l}$ denote the change in layer $l$ after learning task $2$. For $\bm{x}_{i,1} \in \mathcal{S}_{1}^{l}$ from task $1$, it follows that $\Delta \theta_{1}^{l} \bm{x}_{i,1} = 0$ due to the orthogonality of $\bm{g}^{\perp}_{1}$ with respect to $\mathcal{S}_{1}^{l}$ \cite{zhang2021understanding, saha2021gradient}. Therefore, we can obtain:
    \begin{align}
        \theta^{l}_{2} \bm{x}_{i,1}^{l} = (\theta^{l}_{1} + \Delta \theta^{l}_{1}) \bm{x}_{i,1}^{l} = \theta^{l}_{1} \bm{x}_{i,1}^{l}.
    \end{align}
It demonstrates that there is no forgetting of knowledge of task $1$, if the gradient of parameters is orthogonal to the old feature space. We denote the above condition as the {\emph{orthogonal condition}}. 


\textbf{Stage ($2$) Updating of old faeture space.}
After learning task $i$, where $i \geq 2$, $\mathcal{S}_{i-1}^{l}$ in $\mathcal{M}$ needs to be updated to $\mathcal{S}_{i}^{l}$ with new task-wised bases from task $i$. To obtain such bases, for each layer $l$, we need to construct a representation matrix $\bm{R}_{i}^{l}=\left[\bm{x}_{1,1}^{l}, \dots, \bm{x}_{1,n}^{l}\right] \in \mathbb{R}^{n \times d}$ from task $i$ only. Before performing SVD and subsequent $k$-rank approximation, we first eliminate the common bases that already present in $\mathcal{S}_{i-1}^{l}$ so that newly added bases are unique and orthogonal to the existing bases in $\mathcal{S}_{i-1}^{l}$. To accomplish this, we proceed as follows:
    \begin{align}
        {\hat{\bm{R}}}_{i}^{l} = \bm{R}_{i}^{l} - \bm{B}_{i-1}^{l} \left(\bm{B}_{i-1}^{l}\right)^{T}  \left(\bm{R}_{i}^{l}\right) = \bm{R}_{i}^{l} - \bm{R}_{i, \text{proj}}^{l}.
    \end{align}
Afterwards, SVD is performed on ${\hat{\bm{R}}}_{i}^{l}=\hat{\bm{U}}_{i}^{l}\hat{\bm{\Sigma}}_{i}^{l}(\hat{\bm{V}}_{i}^{l})^{T}$, thus obtaining $h$ new orthogonal bases for minimun value of $h$ statisfying the following criteria for the given threshold, $\epsilon_\text{task}$:
    \begin{align}
        ||\bm{R}_{i, \text{proj}}^{l}||_{F}^{2} + ||{\hat{\bm{R}}}_{i}^{l}||_{F}^{2} \geq \epsilon_{\text{task}} ||\bm{R}_{i}^{l}||_{F}^{2}.
    \end{align}
$\bm{B}_{i-1}^{l}$ is then updated to $\bm{B}_{i}^{l}=\left[\bm{B}_{i-1}^{l}, \bm{u}_{1}^{l}, \dots, \bm{u}_{h}^{l}\right]$ with $h$ new bases. And $\mathcal{S}_{i-1}^{l}$ is updated to $\mathcal{S}_{i}^{l}=\text{span}\left\{\bm{B}_{i}^{l}\right\}$. 

\begin{figure*}[t]
    \centering
    \includegraphics[width=1\textwidth]{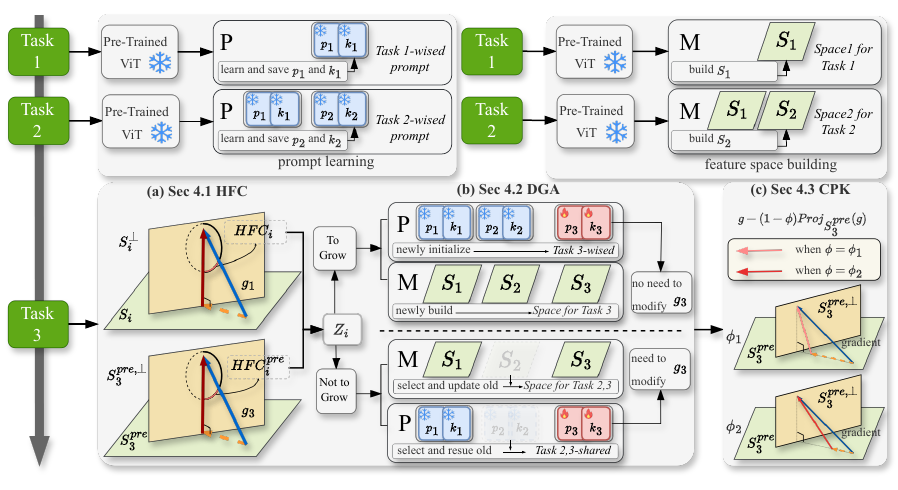}
    \vspace{-3mm}
    \caption{\textbf{Overview of LW2G:} Assume there are already two sets in $\mathcal{P}=\left\{(\bm{p}_1,\bm{k}_1), (\bm{p}_2,\bm{k}_2)\right\}$ and two spaces in $\mathcal{M}=\left\{\mathcal{S}_{1},\mathcal{S}_{2}\right\}$. (a) We use $\mathcal{P}$ and $\mathcal{M}$ to sequentially calculate the hindrance to learning task $3$ on the existing prompt set under the \emph{orthogonal condition} (Sec.\ref{sec_4_1}). (b) Through a dynamic threshold, we decide whether to grow or not to grow. In the former case, $(\bm{p}_3,\bm{k}_3)$ is \textbf{newly initialized} and $\mathcal{S}_{3}$ is \textbf{newly built} and there is no need to modify the original gradient; while in the latter case, $(\bm{p}_3,\bm{k}_3)$ is \textbf{selected from old} $(\bm{p}_2,\bm{k}_2)$ and $\mathcal{S}_{3}$ is \textbf{updated from} $\mathcal{S}_{2}$ ($(\bm{p}_2,\bm{k}_2)$ and $\mathcal{S}_{2}$ will be deleted after learning task $3$) and it is necessary to modify the gradient using the old feature space in $\mathcal{M}$ (Sec.\ref{sec_4_2}). (c) Control the pre-trained knowledge to be consistent with the downstream task knowledge (Sec.\ref{sec_4_3}).}
   \label{main_pic}
\end{figure*}

\paragraph{Prompt-based Continual Learning}
\label{Prompt_extending_CIL_methods_problem}
Recent studies \cite{wang2024hierarchical, smith2023coda, Wang_2022_CVPR, wang2022dualprompt, wang2022s} utilized prompts to leverage the PTMs. Therefore, the DNN is a Vision Transformer (ViT), and the operation of layer $l$, $f^{l}$, is the \emph{attention mechanism} within each transformer block. Hence, the input of ViT after \emph{patch embedding} is $\bm{x}_{e} \in \mathbb{R}^{L_{e} \times d}$, where $L_{e}$ is the token length. Specifically, VPT \cite{jia2022visual, li2021prefix} prepend a set of learnable tokens $\bm{p} \in \mathbb{R}^{L_{p} \times d}$ to $x_{e}$ and treat $\left[\bm{p}, \bm{x}_{e}\right] \in \mathbb{R}^{\left(L_{e}+L_{p}\right) \times d}$ as the input, minimizing $\mathcal{L}_{t}(\mathcal{D}^{t})$ to encode task-wised knowledge into these prompts while keeping pre-trained weights frozen. PCL involves two stages: (a) \emph{prompt learning} and (b) \emph{prompt retrieval}. In \emph{prompt learning}, PCL grows the prompt sets pool $\mathcal{P}$ by initializing a new set of prompt $(\bm{p}_i,\bm{k}_i)$ before learning each new task $i$, where $\bm{p}_i$ is combined with the training samples by the \emph{attention mechanism}. Meanwhile, $\bm{k}_i$ is optimized by being pulled closer to the vanilla features of the training samples obtained by ViT without combining with prompts. In \emph{prompt retrieval}, the vanilla features of each testing sample serve as the query to retrieve from each $\bm{k}_i$ in the prompt sets pool via a carefully designed matching mechanism, such as KNN, cosine similarity, or an auxiliary classifier, and subsequently choose the corresponding $\bm{p}_i$. 

\section{Theory and Method}
\label{sec_4}
In this section, we first present a theoretical analysis of GPCL concerning the hindrance to learning new tasks under the \emph{orthogonal condition} (Theorem \ref{theorem_1} and Definition \ref{define_of_hfc}).   Subsequently, as illustrated in Figure \ref{main_pic}, we simply use an example with three tasks to introduce the plug-in method \textbf{Learning Whether to Grow (LW2G)}, which consists of two components: DGA and CPK. Additionally, a pseudocode for a general scenario is provided in Algorithm \ref{algo_LW2G}.


\subsection{Theoretical Analysis on Hindrance in GPCL}
\label{sec_4_1}

For simplicity, the notation of layer $l$ is omitted in the following analysis. While learning task $i$, GPCL updates the parameters under the \textit{orthogonal condition} to avoid interfering with old knowledge. However, since the gradient represents the direction of local optimal descent for the loss function, modifying it inevitably results in a reduction of local information. To quantify the hindrance from this lost information to learning the new tasks under the \textit{orthogonal condition}, we first define the following metric.

\begin{definition}[Hinder Forward Capability, HFC] 
\label{define_of_hfc}
In GPCL, while continually encoding new knowledge into a single model under the orthogonal condition, Hinder Forward Capability (HFC) is defined to evaluate the hindrance to learning new tasks. HFC is the angle between the original gradient obtained through backpropagation $\bm{g}$ and its projection $\bm{g}^{\perp} = \text{Proj}_{\mathcal{S}^{\perp}_{old}}(\bm{g})$ onto $\mathcal{S}^{\perp}_{old}$, 
        \[\text{HFC}(\bm{g}, \bm{g}^{\perp}) = \arccos\left( \frac{\bm{g} \cdot \bm{g^{\perp}}}{\| \bm{g} \| \| \bm{g}^{\perp} \|} \right).\]
\end{definition}
As illustrated in Figure \ref{gpcl_pic}, a large $\text{HFC}$ indicates a significant gap between original gradient $\bm{g}$ and the modified gradient $\bm{g}^{\perp}$. Therefore, a large reduction of local information leads to greater hindrance to learning new tasks. Based on this, we formally present the following theorem:
\begin{theorem}
\label{theorem_1}
Given a Euclidean space $\mathcal{S}_1 = \text{span} \{\bm{B}_1\}$, where $\bm{B}_1 = [\bm{b}_1, \dots, \bm{b}_l] \in \mathbb{R}^{n \times l}$ is a set of $l$ orthogonal bases for $\mathcal{S}_1$, and a Euclidean space $\mathcal{S}_2 = \text{span} \{\bm{B}_2\}$, where $\bm{B}_2 = [\bm{b}_1, \dots, \bm{b}_l, \bm{b}_{l+1}, \dots, \bm{b}_{l+k}] \in \mathbb{R}^{n \times (l + k)}$ is a set of $l+k$ orthogonal bases for $\mathcal{S}_2$. Then, $\forall \bm{\alpha}$ there always exists:
        \[\text{HFC}(\bm{\alpha}, \text{Proj}_{\mathcal{S}_1}(\bm{\alpha})) > \text{HFC}(\bm{\alpha}, \text{Proj}_{\mathcal{S}_2}(\bm{\alpha})).\]
\end{theorem}

\textit{Proof.} $\forall \bm{\alpha} \in \mathbb{R}^{n \times 1}$, $\bm{\alpha} = \left[\alpha_1, \dots, \alpha_n\right]^{T}$. Without loss of generality, $\left\{\bm{b}_i, i=1,\dots,{k_{1}}+{k_{2}}\right\}$ is a set of \emph{standard orthonormal basis}. As we defined, $\text{Proj}_{\mathcal{S}_1}(\bm{\alpha}) = \left[g_{1}, \dots, g_{{k_{1}}}\right] \in \mathbb{R}^{{k_{1}} \times 1}$ and $\text{Proj}_{\mathcal{S}_2}(\bm{\alpha}) =\left[g_{1}, \dots, g_{{k_{1}}}, g_{{k_{1}}+1}, \dots, g_{{k_{1}}+{k_{2}}}\right]\in \mathbb{R}^{\left({k_{1}} + {k_{2}}\right) \times 1}$, where $g_{i} = \langle \bm{\alpha}, \bm{b_{i}} \rangle$.

Then, we have
\begin{align}
\begin{split}
    cos(\bm{\alpha}, \text{Proj}_{\mathcal{S}_1}(\bm{\alpha})) &= \frac{\bm{\alpha} \cdot \text{Proj}_{\bm{S}_1}(\bm{\alpha})}{\| \bm{\alpha} \| \| \text{Proj}_{\bm{S}_1}(\bm{\alpha}) \|} \\
    &= \frac{\sum_{i=1}^{{k_{1}}}\left(g_{i}\right)^{2}}{\sqrt{\sum_{i=1}^{{k_{1}}}\left(g_{i}\right)^{2}} \sqrt{\sum_{i=1}^{n}\left(g_{i}\right)^{2}}}
    \end{split}
\end{align}

Likewise, we have
\begin{align}
    \begin{split}
        cos(\bm{\alpha}, \text{Proj}_{\bm{S}_2}(\bm{\alpha})) &= \frac{\bm{\alpha} \cdot \text{Proj}_{\bm{S}_2}(\bm{\alpha})}{\| \bm{\alpha} \| \| \text{Proj}_{\bm{S}_2}(\bm{\alpha}) \|} \\
        &= \frac{\sum_{i=1}^{{k_{1}}+{k_{2}}}\left(g_{i}\right)^{2}}{\sqrt{\sum_{i=1}^{{k_{1}}+{k_{2}}}\left(g_{i}\right)^{2}} \sqrt{\sum_{i=1}^{n}\left(g_{i}\right)^{2}}}
        \end{split}
\end{align}

In addition, 
\begin{align}
    \begin{split}
        \frac{cos(\bm{\alpha}, \text{Proj}_{\bm{S}_2}(\bm{\alpha}))}{cos(\bm{\alpha}, \text{Proj}_{\bm{S}_1}(\bm{\alpha}))} &= \frac{\sum_{i=1}^{{k_{1}}+{k_{2}}}\left(g_{i}\right)^{2}}{\sum_{i=1}^{{k_{1}}}\left(g_{i}\right)^{2}} \frac{\sqrt{\sum_{i=1}^{{k_{1}}}\left(g_{i}\right)^{2}}}{\sqrt{\sum_{i=1}^{{k_{1}}+{k_{2}}}\left(g_{i}\right)^{2}}}
        \end{split} \\
        &= \frac{1 + C}{\sqrt{(1 + C)}} \\
        &=\sqrt{(1 + C)} \geq 1.
\end{align}
Where $C=\frac{\sum_{i={k_{1}}+1}^{{k_{1}}+{k_{2}}}\left(g_{i}\right)^{2}}{\sum_{i=1}^{{k_{1}}}\left(g_{i}\right)^{2}} \geq 0$. Thus, we can derive:
\begin{align}
cos(\bm{\alpha}, \text{Proj}_{\bm{S}_2}(\bm{\alpha})) \geq cos(\bm{\alpha}, \text{Proj}_{\bm{S}_1}(\bm{\alpha})).
\end{align}

And, 
\begin{align}
    \text{HFC}(\bm{\alpha}, \text{Proj}_{\bm{S}_1}(\bm{\alpha})) \geq \text{HFC}(\bm{\alpha}, \text{Proj}_{\bm{S}_2}(\bm{\alpha})).
\end{align}

This finishes the proof.

The above Theorem \ref{theorem_1} shows that fewer bases result in a larger HFC value. As old feature space , $\mathcal{S}_{old}$, in $\mathcal{M}$ continues to expand with new orthogonal bases from each new task, its corresponding orthogonal complement space, $\mathcal{S}_{old}^{\perp}$, progressively shrinks. Consequently, the bases in $\mathcal{S}_{old}^{\perp}$ steadily decrease, leading to a large HFC value and more severe hindrance to learning new tasks under the \emph{orthogonal condition}.
\vspace{-5mm}
\subsection{Dynamic Growing Approach}
\label{sec_4_2}
Instead of naively growing a new set of prompts for each new task regardless of task dissimilarities, we propose a \textbf{Dynamic Growing Approach (DGA)}. DGA involves dynamically learning whether \emph{to grow} (initialize a new set of prompts and store it in the pool) or \emph{not to grow} (utilize an existing set from the pool). 

Before learning task $3$, we first qualify the hindrance to learning task $3$ on each old set in the prompt pool under the \emph{orthogonal condition}. Specifically, we iteratively select an \textbf{old} set $(\bm{p}_1,\bm{k}_1)$ from $\mathcal{P}$ and an \textbf{old} space $\mathcal{S}_{1}$ from $\mathcal{M}$, where $\mathcal{S}_{1}=\left\{\mathcal{S}_{1}^{l}\right\}_{l \in L_{\text{act}}}$ is the feature space corresponding to task $1$ and $L_{\text{act}}$ is the set of transformer blocks on which the prompt inserts in the baseline method\footnote{For example, in DualPrompt, prompt is inserted on the first 5 transformer blocks, then $L_{\text{act}} = \left\{1,2,3,4,5\right\}$.}. We construct a subset of training dataset from task $3$, denoted as $\mathcal{D}^{3}_{\text{sub}}$. For clarity, the gradient to update $(\bm{p}_1,\bm{k}_1)$ with $\mathcal{D}^{3}_{\text{sub}}$ is denoted as:
\begin{align}
    \label{gradient_1_calculate}
    \bm{g}_{1} = \nabla_{(\bm{p}_1,\bm{k}_1)} \mathcal{L}_{3}(\mathcal{D}^{3}_{\text{sub}}).
\end{align}    
To prevent the influence of old knowledge contained in $(\bm{p}_1,\bm{k}_1)$ while learning task $3$, the gradient $\bm{g}_{1}$ is required to be modified to $\text{Proj}_{\mathcal{S}^{\perp}_{1}}(\bm{g}_{1})$, where $\mathcal{S}^{\perp}_{1}$ is the orthogonal complement of $\mathcal{S}_{1}$. Then, $\text{Proj}_{\mathcal{S}^{\perp}_{1}}(\bm{g}_{1})$ serves as the real gradient for updating parameters. Based on Theorem \ref{theorem_1}, we evaluate the hindrance under the \emph{orthogonal condition} while learning task $3$ on $(\bm{p}_1,\bm{k}_1)$ as follows:
\begin{align}
    \label{tab:modified_g_with_old_task}
    \text{HFC}_1 = \text{HFC}(\bm{g}_{1}, \text{Proj}_{\mathcal{S}^{\perp}_{1}}(\bm{g}_{1})).
\end{align}
Besides, we define a dynamic threshold based on the task $3$ and the PTM being used. Firstly, we \textbf{temporarily} initialize a \textbf{new} set with $(\bm{p}_1,\bm{k}_1)$ as follows:
\begin{align}
\label{initial}
    (\bm{p}_3,\bm{k}_3) \Leftarrow (\bm{p}_1,\bm{k}_1).
\end{align}
Here, as the newly initialized $(\bm{p}_3,\bm{k}_3)$ is dedicated to learning task $3$ and does not need to serve as a shared prompt set with any old task (task $1$ or task $2$), it represents an ideal scenario for learning task $3$. Likewise, the gradient to update $(\bm{p}_3,\bm{k}_3)$ is denoted as:
\begin{align}
\label{gradient_2_calculate}
    \bm{g}_{3} = \nabla_{(\bm{p}_3,\bm{k}_3)} \mathcal{L}_{3}(\mathcal{D}^{3}_{\text{sub}}).
\end{align}    
Then, we can obtain a valinia representation matrix $\bm{R}_{3}^{\text{pre}}$ by feeding $\mathcal{D}^{3}_{\text{sub}}$ into the ViT without prompts. We can newly build $\mathcal{S}_{3}^{\text{pre}}$ after performing SVD and $k$-rank approximation with pre-trained threshold, $\epsilon_\text{pre}$. Then, we can also calculate: 
\begin{align}
    \label{tab:modified_g_with_pre_task}
    \text{HFC}_{1}^{\text{pre}}=\text{HFC}(\bm{g}_{3}, \text{Proj}_{\mathcal{S}_{3}^{\text{pre}, \perp}}(\bm{g}_{3})),
\end{align}
where $\mathcal{S}_{3}^{\text{pre}, \perp}$ is the orthogonal complement of $\mathcal{S}_{3}^{\text{pre}}$.
Here, $\text{HFC}_{1}^{\text{pre}}$ represents the relationship between the gradient of learning task $3$ and the pre-trained knowledge from task $3$. As we initialize $(\bm{p}_3,\bm{k}_3)$ with $(\bm{p}_1,\bm{k}_1)$ via Eq.(\ref{initial}) to ensure that the differences in HFC values (Eq.(\ref{tab:modified_g_with_old_task}) v.s. Eq.(\ref{tab:modified_g_with_pre_task})) arise solely from the difference of feature projection spaces ($S_{1}^{\perp}$ v.s. $S_{3}^{\text{pre},\perp}$), rather than from the gradient magnitudes. Therefore, $\text{HFC}_{1}^{\text{pre}}$ signifies the ideal scenario when learning new tasks in PCL, which is the \emph{dynamic threshold to evaluate the relative magnitude of hindrance}. Based on this, the gap between learning on \textbf{old} set $(\bm{p}_1,\bm{k}_1)$ under the \emph{orthogonal condition} and leaning on \textbf{new} set $(\bm{p}_3,\bm{k}_3)$ in an ideal scenario with no gradient modification is denoted as follows:
\begin{align}
    \label{tab:get_z}
    Z_{1} =  \text{HFC}_{1} - \text{HFC}_{1}^{\text{pre}}.
\end{align}
Thus, if $Z_{1}>0$, it indicates that learning task $3$ on the old set $(\bm{p}_1, \bm{k}_1)$ from $\mathcal{P}$ under the \textit{orthogonal condition} suffers hindrance ($HFC_{1}$) compared with the ideal scenario where task $3$ is learned on newly initialized $(\bm{p}_3, \bm{k}_3)$ without any constraints ($HFC_{1}^{\text{pre}}$). Furthermore, the magnitude of $Z_{1}$ indicates the degree of the hindrance, with a large value implying excessive hindrance. Likewise, the gap between learning on \textbf{old} set $(\bm{p}_2,\bm{k}_2)$ under the \emph{orthogonal condition} and leaning on \textbf{new} set $(\bm{p}_3,\bm{k}_3)$ \textbf{temporarily} initialized with $(\bm{p}_2,\bm{k}_2)$ in an ideal scenario can also be calculated as $Z_{2}$.

\paragraph{Opting To Grow or Not To Grow}
Based on the analysis, we propose a dynamic growing approach as follows:
    \begin{align}
        \label{tab:decide}
        \left\{
    \begin{array}{ll}
        \emph{  \, \, \,  \,  To Grow} \, \, \,\,\, \quad \quad \text{if} \quad  &\underset{{m \in (1,2)}}{\min} Z_{m} > 0 \\
        \emph{Not To Grow} \quad \quad \quad \text{else}   \quad &\underset{{m \in (1,2)}}{\min} Z_{m} \leq 0.
    \end{array}
        \right.
    \end{align}
\setlength{\leftmargini}{8pt} 
\begin{itemize}
    \item While chosing \textbf{To Grow}, we permanently initialize a new set $(\bm{p}_3,\bm{k}_3)$. Then, update $(\bm{p}_3,\bm{k}_3)$ with task $3$ and build a new feature space $\mathcal{S}_{3}$ with threshold, $\epsilon_{\text{task}}$, from task $3$ only and store $\mathcal{S}_{3}$ into $\mathcal{M}$.
    \item While chosing \textbf{Not To Grow}, we select an old set $(\bm{p}_j,\bm{k}_j)$ from $\mathcal{P}$, where $j = {\arg\min}_{{m \in (1,2)}} {Z_{m}}$.  Then, update $(\bm{p}_j,\bm{k}_j)$ with task $3$ under \emph{orthogonal condition} and update the old feature space $\mathcal{S}_{j}$ with threshold, $\epsilon_{\text{task}}$, with new bases from task $3$.
\end{itemize}
Details about building of new feature or updating of old feature space are shown in Section \ref{pre_gpcl_methods}.


\subsection{Consistency with Pre-trained Knowledge}
\label{sec_4_3}
Recent studies in transfer learning and domain adaptation revealed that when employing PEFT for fine-tuning PTM, the performance after fine-tuning often falls short of the pre-trained knowledge of PTM itself. For example, \cite{lee2023read, zhu2023prompt, khattak2023self} have shown that even when freezing the backbone and using PEFT, e.g., prompt or LoRA, it still cannot fully guarantee the preservation of pre-trained knowledge. However, previous works only focus on the catastrophic forgetting of old tasks during the continual learning of new tasks, while ignoring the fact that continuous fine-tuning on downstream tasks can also lead to the forgetting of pre-trained knowledge. This oversight may gradually affect the learning of new tasks. For example, in DualPrompt \cite{wang2022dualprompt}, the $g$-prompt is updated across all tasks, which can influence the underlying representations. In CODA-Prompt \cite{smith2023coda} and OSPrompt \cite{kim2025one}, prompt sets belong to old tasks still contribute to the training of the new prompt set. Similarly, in LoRA-based methods, each LoRA module is merged into the backbone aftering learning old tasks. These operations, although not directly modifying the backbone, can still affect how the pre-trained knowledge is utilized and may hinder the learning of new tasks.

Based on these, we exploit two distinct levels of forgetting issues faced in PCL: (1) continuous fine-tuning on downstream tasks leading to the forgetting of pre-trained knowledge, and (2) continual learning on new tasks resulting in the forgetting of old tasks.

To tackle the former issue, we adjust the gradient of the new tasks to be orthogonal to the pre-trained feature space. However, due to the domain gap between the incremental task training data and the pre-trained data, a fully orthogonal constraint is too stringent and can significantly impact plasticity. To achieve a balance between maintaining plasticity and fully utilizing the pre-trained knowledge, we propose to apply a soft constraint as follows:
\begin{align}
    \label{tab:soft_constraints_on_g}
    \bm{g} = \bm{g} - (1-\phi) \text{Proj}_{\mathcal{S}^{\text{pre}}_{3}}(\bm{g}),
\end{align}
where $\phi$ is the coefficient of the soft constraint to control the orthogonality and $\mathcal{S}_{3}^{\text{pre}}$ is the pre-trained feature space for task $3$. When learning on task $3$, the gradient can be obtained from Eq.(\ref{gradient_1_calculate}) while DGA chooses \textit{to grow}, or from Eq.(\ref{gradient_2_calculate}) while DGA chooses \textit{not to grow}. And $\phi$ can flexibly control the real gradient $\bm{g}$, aligning it as closely as possible with the feature space of the pre-trained knowledge, while ensuring the learning ability on new tasks.

\begin{algorithm}[h]
	\small
	\caption{LW2G: Learning Whether to Grow.}
	\label{algo_LW2G}
	\raggedright
	{\bf Input}: Task length $T$, Datasets for each task: $\{\mathcal{D}^{1}, \mathcal{D}^{2}, \cdots, \}$, Pool $\mathcal{P}=\left\{\right\}$, Memory $\mathcal{M}=\left\{\right\}$, Training Epochs $E$. \\
	{\bf Output}: Updated Pool $\mathcal{P}$ and Memory $\mathcal{M}$.
	\begin{algorithmic}[1]
        \For{$i = 1, 2, \cdots, T$}
            \Comment{{\textbf{DGA}}}
            \If{$i$ = 1} 
                \State \textbf{DGA} chose to grow;
                \State Initialization $\left(p_{i},k_{i}\right)$ and Store it in $\mathcal{P}$;
            \Else
                \State Get a subset from $\mathcal{D}^{i}_{\text{sub}}$.
                \State Get all selectable sets in $\mathcal{P}$, denoted as L;
                \For{$\text{j} \, \,  \text{in} \, \, L$} 
                    \State Get an old set from $\mathcal{P}$, as $\left(p_{j},k_{j}\right)$;
                    \State Get an old feature space from $\mathcal{M}$, as $\mathcal{S}_{j}$;
                    \State Get $\bm{g}$ on $\left(p_{j},k_{j}\right)$ with $\mathcal{D}^{i}_{\text{sub}}$;
                    \State Get $\text{HFC}_{\text{j}}$ via Eq.(\ref{tab:modified_g_with_old_task});
                    \State Get $\text{HFC}_{\text{pre}}$ via Eq.(\ref{tab:modified_g_with_pre_task});
                    \State Get $Z_{\text{j}}$ via Eq.(\ref{tab:get_z}); \Comment{{HFC}};
                \EndFor
                \If{\textbf{DGA} chose to grow}
                    \State Initialization $\left(p_{i},k_{i}\right)$ and Store it in $\mathcal{P}$;
                \Else     
                    \State Selection $\left(p_{t},k_{t}\right)$, $t = {\arg\max}_{{j \in L}} {Z_{j}}$;
                    \State Change $\left(p_{t},k_{t}\right)$ to $\left(p_{i},k_{i}\right)$;
                    \State Change $\mathcal{S}_{t}$ to $\mathcal{S}_{i}$;
                \EndIf      
            \EndIf
            \For{$e = 1, 2, \cdots, E$} \Comment{{\textbf{Start Training}}}
                \State Get $\bm{g}$ on $\left(p_{i},k_{i}\right)$ with $\mathcal{D}^{i}$;
                \State Modify $\bm{g}$ via Eq.(\ref{tab:soft_constraints_on_g}); \Comment{{\textbf{CPK}}}
                \State Update $\left(p_{i},k_{i}\right)$;
                \EndFor
            \State Build or update space $\mathcal{S}_{i}$ in $\mathcal{M}$ via Section \ref{pre_gpcl_methods};
            \EndFor

		\Return $\mathcal{P}$, $\mathcal{M}$;
	\end{algorithmic}
\end{algorithm}

\begin{table*}[t!]
  \centering
  \caption{We compare LW2G with state-of-the-art (a) single-set methods, (b) prompt-based methods, and (c) LoRA-based methods. Notably, for each metric, $\uparrow$ ($\downarrow$) indicates that the larger (the smaller) values, the better results are. OnePrompt, Aper-VPT-Deep, and Aper-Adapter use only a single prompt set; CODA-Prompt and OSPrompt apply a weighting mechanism over all prompt sets; C-LoRA and InfLoRA-b5 utilize a LoRA module with a merging mechanism. They all bypass the \textit{prompt retrieval} stage. Therefore, we do not provide the PRA metric for these baselines. The best results are highlighted in bold.}
\label{tab:main_table1}
    \vspace{-0.32cm}
  \resizebox{1\textwidth}{!}{
  \begin{tabular}
  {p{2.5cm}p{2.0cm}p{1.5cm}p{1.5cm}p{1.3cm}p{1.2cm}p{0.cm}p{1.5cm}p{1.5cm}p{1.3cm}p{1.2cm}}
  \toprule
   \multirow{3}{*}{Method} & \multirow{3}{*}{Publication} & \multicolumn{4}{c}{\emph{CIFAR(Inc10Task10)}} && \multicolumn{4}{c}{\emph{IMR(Inc20Task10)}} \\
  \cmidrule{3-6} \cmidrule{8-11}
   & & FAA ($\uparrow$) & PRA ($\uparrow$) & FFM ($\downarrow$) & SSP ($\downarrow$) && FAA ($\uparrow$) & PRA ($\uparrow$) & FFM ($\downarrow$) & SSP ($\downarrow$) \\
    \midrule
    \multicolumn{5}{l}{\textcolor{orange}{\textit{(a) single-prompt methods}}} \\
     OnePrompt & \multirow{1}{*}{ICLR'24} & 85.60 & - & 4.79 & 1 && 71.34 & - & 7.36 & 1 \\
     Aper-VPT-Deep & \multirow{1}{*}{IJCV'25} & 84.95 & - & 6.55 & 1 && 69.47 & - & 7.07 & 1 \\
     Aper-Adapter & \multirow{1}{*}{IJCV'25} & 85.68 & - & 5.78 & 1 && 67.95 & - & 6.92 & 1 \\
    \midrule
    \multicolumn{5}{l}{\textcolor{orange}{\textit{(b) LW2G plug-in with prompt-based methods}}} \\
    DualPrompt & \multirow{1}{*}{ECCV'22} & 85.94\scriptsize{$\pm$0.1} & 59.44\scriptsize{$\pm$0.3} & 6.38\scriptsize{$\pm$0.1} & 10 && 67.30\scriptsize{$\pm$0.3} & 45.51\scriptsize{$\pm$0.9} & 6.41\scriptsize{$\pm$0.1} & 10 \\
    \rowcolor{LightCyan}\quad w/ LW2G && \textbf{86.86}\scriptsize{$\pm$0.3} & \textbf{78.33}\scriptsize{$\pm$0.1} & \textbf{6.03}\scriptsize{$\pm$0.6} & \textbf{2} && \textbf{68.68}\scriptsize{$\pm$0.5} & \textbf{78.90}\scriptsize{$\pm$1.3} & \textbf{5.72}\scriptsize{$\pm$0.1} & \textbf{2} \\
    && \gr{+0.92} & \gr{+18.89} & \bluegr{-0.35} & \bluegr{-8} && \gr{+1.38} & \gr{+33.39} & \bluegr{-0.69} & \bluegr{-8} \\

     S-Prompt++ & \multirow{1}{*}{NeurIPS'24} & 89.25\scriptsize{$\pm$0.0} & 99.52\scriptsize{$\pm$0.5} & 4.10\scriptsize{$\pm$0.2} & 10 && 66.03\scriptsize{$\pm$0.3} & 45.73\scriptsize{$\pm$0.6} & 6.22\scriptsize{$\pm$0.2} & 10 \\
     \rowcolor{LightCyan}\quad w/ LW2G && \textbf{89.32}\scriptsize{$\pm$0.0} & \textbf{100.0}\scriptsize{$\pm$0.0} & \textbf{3.46}\scriptsize{$\pm$0.4} & \textbf{7} && \textbf{68.18}\scriptsize{$\pm$0.3} & \textbf{78.80}\scriptsize{$\pm$0.3} & \textbf{6.01}\scriptsize{$\pm$0.1} & \textbf{5} \\
      && \gr{+0.07} & \gr{+0.48} & \bluegr{-0.64} & \bluegr{-3} && \gr{+2.15} & \gr{+33.07} & \bluegr{-0.21} & \bluegr{-5} \\

      HiDe-Prompt & \multirow{1}{*}{NeurIPS'24} & 85.77\scriptsize{$\pm$0.1} & 80.78\scriptsize{$\pm$0.1} & 6.19\scriptsize{$\pm$0.1} & 10 && 65.07\scriptsize{$\pm$0.3} & 64.00\scriptsize{$\pm$0.2} & 8.89\scriptsize{$\pm$0.1} & 10 \\
      \rowcolor{LightCyan}\quad w/ LW2G && \textbf{87.60}\scriptsize{$\pm$0.1} & \textbf{95.39}\scriptsize{$\pm$0.2} & \textbf{4.28}\scriptsize{$\pm$0.3} & \textbf{2} && \textbf{65.84}\scriptsize{$\pm$0.3} & \textbf{66.78}\scriptsize{$\pm$0.9} & \textbf{7.19}\scriptsize{$\pm$0.3} & \textbf{6} \\
       && \gr{+1.83} & \gr{+14.61} & \bluegr{-1.91} & \bluegr{-8} && \gr{+0.77} & \gr{+2.78} & \bluegr{-1.70} & \bluegr{-2} \\

       CPrompt & \multirow{1}{*}{CVPR'24} & 86.13\scriptsize{$\pm$0.0} & 69.28\scriptsize{$\pm$0.2} & 6.00\scriptsize{$\pm$0.1} & 10 && 74.83\scriptsize{$\pm$0.2} & 63.20\scriptsize{$\pm$0.3} & 7.26\scriptsize{$\pm$0.2} & 10 \\
       \rowcolor{LightCyan}\quad w/ LW2G && \textbf{86.93}\scriptsize{$\pm$0.0} & \textbf{80.17}\scriptsize{$\pm$0.2} & \textbf{4.72}\scriptsize{$\pm$0.3} & \textbf{5} && \textbf{76.85}\scriptsize{$\pm$0.6} & \textbf{81.01}\scriptsize{$\pm$0.4} & \textbf{6.37}\scriptsize{$\pm$0.1} & \textbf{2} \\
        && \gr{+0.80} & \gr{+10.89} & \bluegr{-1.28} & \bluegr{-5} && \gr{+2.02} & \gr{+17.81} & \bluegr{-0.89} & \bluegr{-8} \\

        CODA-Prompt & \multirow{1}{*}{CVPR'23} & 86.72\scriptsize{$\pm$0.1} & - & 4.04\scriptsize{$\pm$0.1} & 10 && 75.73\scriptsize{$\pm$0.1} & - & 5.17\scriptsize{$\pm$0.2} & 10 \\
        \rowcolor{LightCyan}\quad w/ LW2G && \textbf{87.33}\scriptsize{$\pm$0.2} & - & \textbf{3.81}\scriptsize{$\pm$0.1} & \textbf{3} && \textbf{76.63}\scriptsize{$\pm$0.2} & - & \textbf{4.39}\scriptsize{$\pm$0.1} & \textbf{3} \\
         && \gr{+0.61} & - & \bluegr{-0.23} & \bluegr{-7} && \gr{+0.90} & - & \bluegr{-0.78} & \bluegr{-7} \\

         OSPrompt & \multirow{1}{*}{ECCV'24} & 86.96\scriptsize{$\pm$0.1} & - & 3.90\scriptsize{$\pm$0.1} & 10 && 75.55\scriptsize{$\pm$0.1} & - & 5.36\scriptsize{$\pm$0.2} & 10 \\
         \rowcolor{LightCyan}\quad w/ LW2G && \textbf{87.59}\scriptsize{$\pm$0.2} & - & \textbf{3.53}\scriptsize{$\pm$0.5} & \textbf{3} && \textbf{76.13}\scriptsize{$\pm$0.2} & - & \textbf{4.61}\scriptsize{$\pm$0.1} & \textbf{3} \\
          && \gr{+0.63} & - & \bluegr{-0.37} & \bluegr{-7} && \gr{+0.58} & - & \bluegr{-0.75} & \bluegr{-7} \\

    \midrule
    \multicolumn{5}{l}{\textcolor{orange}{\textit{(c) LW2G plug-in with LoRA-based methods}}} \\                      
    C-LoRA & \multirow{1}{*}{TMLR'24} & 82.97\scriptsize{$\pm$0.1} & - & 6.73\scriptsize{$\pm$0.1} & 10 && 71.95\scriptsize{$\pm$0.3} & - & 5.82\scriptsize{$\pm$0.1} & 10 \\
    \rowcolor{LightCyan}\quad w/ LW2G && \textbf{84.69}\scriptsize{$\pm$0.3} & - & \textbf{6.24}\scriptsize{$\pm$0.2} & \textbf{2} && \textbf{72.69}\scriptsize{$\pm$0.2} & - & \textbf{5.71}\scriptsize{$\pm$0.0} & \textbf{2} \\
     && \gr{+1.72} & - & \bluegr{-0.49} & \bluegr{-8} && \gr{+1.74} & - & \bluegr{-0.11} & \bluegr{-8} \\

     InfLoRA-b5 & \multirow{1}{*}{CVPR'24} & 86.65\scriptsize{$\pm$0.0} & - & 6.22\scriptsize{$\pm$0.1} & 10 && 73.05\scriptsize{$\pm$0.1} & - & 5.73\scriptsize{$\pm$0.1} & 10 \\
     \rowcolor{LightCyan}\quad w/ LW2G && \textbf{86.81}\scriptsize{$\pm$0.2} & - & \textbf{6.03}\scriptsize{$\pm$0.4} & \textbf{2} && \textbf{73.32}\scriptsize{$\pm$0.4} & - & \textbf{4.93}\scriptsize{$\pm$0.2} & \textbf{3} \\
      && \gr{+0.16} & - & \bluegr{-0.19} & \bluegr{-8} && \gr{+0.27} & - & \bluegr{-0.80} & \bluegr{-7} \\

      HiDe-LoRA & \multirow{1}{*}{TPAMI'25} & 91.21\scriptsize{$\pm$0.0} & 81.60\scriptsize{$\pm$0.1} & 3.36\scriptsize{$\pm$0.2} & 10 && 78.86\scriptsize{$\pm$0.7} & 65.13\scriptsize{$\pm$0.4} & 2.07\scriptsize{$\pm$0.2} & 10 \\
      \rowcolor{LightCyan}\quad w/ LW2G && \textbf{92.89}\scriptsize{$\pm$0.4} & \textbf{95.30}\scriptsize{$\pm$0.1} & \textbf{2.97}\scriptsize{$\pm$0.4} & \textbf{4} && \textbf{79.65}\scriptsize{$\pm$0.4} & \textbf{89.64}\scriptsize{$\pm$0.4} & \textbf{1.85}\scriptsize{$\pm$0.4} & \textbf{5} \\
       && \gr{+1.68} & \gr{+13.70} & \bluegr{-0.39} & \bluegr{-6} && \gr{+0.79} & \gr{+24.51} & \bluegr{-0.22} & \bluegr{-5} \\
    
    \midrule
    \multicolumn{5}{l}{\textcolor{orange}{\textit{Overall average improvement}}} \\
    \multicolumn{2}{l}{\textcolor{black}{Baseline w/ LW2G}} & \cellcolor{LightCyan}\textbf{+0.95} & \cellcolor{LightCyan}\textbf{+11.74} & \cellcolor{LightCyan}\textbf{-0.65} & \cellcolor{LightCyan}\textbf{-6.67} &\cellcolor{LightCyan}& \cellcolor{LightCyan}\textbf{+1.07} & \cellcolor{LightCyan}\textbf{+12.40} & \cellcolor{LightCyan}\textbf{-0.68} & \cellcolor{LightCyan}\textbf{-6.56} \\
    
  \bottomrule
\end{tabular}
}
\end{table*}

\begin{table*}[t]
  \centering
  \caption{We compare LW2G with two existing plug-in methods, LGCL and PGP with two baselines, DualPrompt and S-Prompt++.}
\label{tab:main_table2}
  \resizebox{1\textwidth}{!}{
  \begin{tabular}
  {p{2.5cm}p{1.5cm}p{1.5cm}p{1.3cm}p{1.3cm}p{0.cm}p{1.5cm}p{1.5cm}p{1.3cm}p{1.3cm}}
  \toprule
   \multirow{3}{*}{Method} & \multicolumn{4}{c}{\emph{OMNI(Inc10Task30)}} && \multicolumn{4}{c}{\emph{OMNI(Inc5Task60)}} \\
  \cmidrule{3-5} \cmidrule{7-10}
   & FAA ($\uparrow$) & PRA ($\uparrow$) & FFM ($\downarrow$) & SSP ($\downarrow$) && FAA ($\uparrow$) & PRA ($\uparrow$) & FFM ($\downarrow$) & SSP ($\downarrow$) \\
    \midrule
    DualPrompt &  
            63.36 &
            68.47 &
            12.92 &
            30 &&
            61.85 &
            69.94 &
            13.50 &
            60 \\
    \quad w/ LGCL &             
            64.19 &
            68.47 &
            11.05 &
            30 &&
            62.30 &
            69.94 &
            13.50 &
            60 \\
    \quad w/ PGP & 
            63.74 &
            67.95 &
            12.97 & 
            30 &&
            62.24 &
            68.68 &
            14.64 &
            60 \\
    \rowcolor{LightCyan}\quad w/ LW2G & 
        \textbf{65.12} &
        \textbf{80.95} &
        \textbf{10.75} &
        \textbf{9} &&
        \textbf{63.17} &
        \textbf{75.31} &
        \textbf{12.01} &
        \textbf{17} \\
     & \gr{+1.76} & \gr{+12.21} & \bluegr{-2.17} & \bluegr{-21} && \gr{+1.32} & \gr{+5.37} & \bluegr{-1.49} & \bluegr{-43} \\
      \midrule
    S-Prompt++ &  
            64.44 &
            55.87 &
            9.02 &
            30 &&
            62.31 &
            54.59 &
            10.04 &
            60 \\
    \quad w/ LGCL &             
            64.93 &
            55.87 &
            8.79 &
            30 &&
            62.68 &
            54.59 &
            11.56 &
            60 \\
    \quad w/ PGP & 
            64.40 &
            55.87 &
            9.13 &
            30 &&
            63.01 &
            54.59 &
            10.00 &
            60 \\
    \rowcolor{LightCyan}\quad w/ LW2G & 
            \textbf{65.90} &
            \textbf{63.86} &
            \textbf{8.50} &
            \textbf{10} &&
            \textbf{63.70} &
            \textbf{62.60} &
            \textbf{9.90} &
            \textbf{18} \\
     & \gr{+1.46} & \gr{+7.99} & \bluegr{-0.52} & \bluegr{-20} && \gr{+1.39} & \gr{+8.01} & \bluegr{-0.14} & \bluegr{-42} \\
      \bottomrule
\end{tabular}
}
\end{table*}

\begin{table*}[t!]
  \centering
    \caption{Results on the OMNI with two extreme settings: 30 tasks and 60 tasks. Additionally, we provide results of (a) overhead, including prompt denoted as the total learnable prompts, and base denoted as the orthogonal bases obtained from SVD; (b) computation burden, including FLOPs and Running Time(R.T.). Note that both the learnable prompts and orthogonal bases are stored as a batch of tensors with a length of 768 in float32 format.}
\label{tab:long_task_table_overheads}
  \resizebox{1\textwidth}{!}{
  \begin{tabular}{lccccccccccccccccc}
  \toprule
   \multirow{3}{*}{Method} & \multicolumn{8}{c}{\emph{OMNI(Inc10Task30)}} && \multicolumn{8}{c}{\emph{OMNI(Inc5Task60)}} \\
   \cmidrule{2-9} \cmidrule{11-18} 
   & \multicolumn{1}{c}{\emph{performance ($\uparrow$)}} && \multicolumn{3}{c}{\emph{overhead ($\downarrow$)}} && \multicolumn{2}{c}{\emph{calculation burden ($\downarrow$)}} && \multicolumn{1}{c}{\emph{performance ($\uparrow$)}} && \multicolumn{3}{c}{\emph{overhead ($\downarrow$)}} && \multicolumn{2}{c}{\emph{calculation burden ($\downarrow$)}} \\
   \cmidrule{2-2}  \cmidrule{4-6} \cmidrule{8-9} \cmidrule{11-11} \cmidrule{13-15} \cmidrule{17-18}   
   & FAA  && \multicolumn{1}{c}{prompt} & \multicolumn{1}{c}{base}  & \multicolumn{1}{c}{Overall} && \multicolumn{1}{c}{FLOPs(G)} & \multicolumn{1}{c}{R.T.(h)} && FAA && \multicolumn{1}{c}{prompt}  & \multicolumn{1}{c}{base}  & \multicolumn{1}{c}{Overall} && \multicolumn{1}{c}{FLOPs(G)} & \multicolumn{1}{c}{R.T.(h)} \\
    \midrule
    DualPrompt & 63.36 && \multicolumn{1}{c}{3600} & \multicolumn{1}{c}{0} & \multicolumn{1}{c}{3600} && 35.19 & 4.5 && 61.85 && \multicolumn{1}{c}{7200} & \multicolumn{1}{c}{0} & \multicolumn{1}{c}{7200} && 35.19 & 5.0 \\
    \rowcolor{LightCyan}\quad w/ LW2G & \textbf{65.12} && \multicolumn{1}{c}{1080} & \multicolumn{1}{c}{429} & \multicolumn{1}{c}{1509} && 37.21 & 5.0 && \textbf{63.17} && \multicolumn{1}{c}{2040} & \multicolumn{1}{c}{576} & \multicolumn{1}{c}{2616} && 37.21 & 6.1 \\
     & \gr{+1.76} && \multicolumn{3}{c}{\cellcolor{MistyRose}\bluegr{-2091 $\times$ 768} $\approx$ \textbf{-1.6M}} && \gr{+5.0\%} & \gr{+11.0\%} && \gr{+1.32} && \multicolumn{3}{c}{\cellcolor{MistyRose}\bluegr{-4584 $\times$ 768} $\approx$ \textbf{-3.5M}} && \gr{+5\%} & \gr{+22.0\%} \\
    S-Prompt++ & 64.44 && \multicolumn{1}{c}{6000} & \multicolumn{1}{c}{0} & \multicolumn{1}{c}{6000} && 35.17 & 4.5 && 62.31 && \multicolumn{1}{c}{12000} & \multicolumn{1}{c}{0} & \multicolumn{1}{c}{12000} && 35.17 & 5.1 \\
    \rowcolor{LightCyan}\quad w/ LW2G & \textbf{65.90} && \multicolumn{1}{c}{2000} & \multicolumn{1}{c}{509} & \multicolumn{1}{c}{2509} && 37.24 & 5.2 && \textbf{63.70} && \multicolumn{1}{c}{3600} & \multicolumn{1}{c}{640} & \multicolumn{1}{c}{4240} && 37.24 & 6.2 \\
     & \gr{+1.46} && \multicolumn{3}{c}{\cellcolor{MistyRose}\bluegr{-3771 $\times$ 768} $\approx$ \textbf{-3M}} && \gr{+6.0\%} & \gr{+16.0\%} && \gr{+1.39} && \multicolumn{3}{c}{\cellcolor{MistyRose}\bluegr{-8024 $\times$ 768} $\approx$ \textbf{-6M}} && \gr{+6.0\%} & \gr{+21.0\%} \\

  \bottomrule
\end{tabular}
}
\end{table*}

\section{Experiment}
In this section, we first describe the experimental setups, and then present the experimental results.
\subsection{Experimental Setups}
\paragraph{Benchmark} We evaluate LW2G on multiple datasets against state-of-the-art baselines. Specifically, we use the following datasets: CIFAR100 \cite{krizhevsky2009learning} (CIFAR), ImageNet-R \cite{hendrycks2021many} (IMR), and Omnibenchmark \cite{zhang2022benchmarking} (OMNI). Besides, we denote different experimental settings as `$\text{Dataset(IncNTaskN)}$', e.g., `$\text{CIFAR(INC10Task10)}$', which means learning on CIFAR with 10 tasks and each task contains 10 classes. 
\paragraph{Baseline} We adopt 14 of the current state-of-the-art continual learning methods that utilize a pre-trained model as our baseline. Specifically, we categorize these methods into: (a) single-set methods: OnePrompt \cite{huang2024ovor}, Aper-VPT-Deep and Aper-Adapter \cite{zhou2025revisiting} (b) prompt-based methods: DualPrompt \cite{wang2022dualprompt}, S-Prompt++ \cite{wang2024hierarchical}, Hide- \cite{wang2024hierarchical}, CPrompt \cite{gao2024consistent}, CODA-Prompt \cite{smith2023coda}, and OSPrompt \cite{kim2025one}; (c) LoRA-based methods: C-LoRA \cite{smith2023continual}, InfLoRA-b5 \cite{liang2024infLoRA}, and HiDe-LoRA \cite{wang2024hide}; (d) other plug-in methods: LGCL \cite{khan2023introducing} and PGP \cite{qiao2023prompt}. 

\paragraph{Implementation} Following \cite{wang2024hide}, we adopt a variety of pre-training paradigms for ImageNet-21K and ImageNet-1K, including Sup-21K, iBOT-21K, iBOT-1K, and DINO-1K. The main results reported in Tables \ref{tab:main_table1}, \ref{tab:main_table2}, \ref{tab:long_task_table_overheads}, \ref{tab:ablation_table}, and \ref{tab:variants_on_dsa} are obtained under Sup-21K, and additional results with other pre-training paradigms are provided in Table \ref{tab:appendix_table_ibot21k}. \textbf{Moreover, to ensure a fair comparison, we implement all baselines using their officially released code and keep their original hyperparameters unchanged.}


\paragraph{Metric}
Following \cite{wang2024hierarchical}, we record the average accuracy over all encountered classes after training on each task. The \textbf{Final Average Accuracy (FAA)} and the \textbf{Final Forgetting Measure (FFM)} are defined based on these results as follows:

\begin{align}
    \text{FAA} &= \frac{1}{T} \sum_{i=1}^{T} A_{i,T}, \\
    \text{FFM} &= \frac{1}{T-1} \sum_{i=1}^{T-1} \max_{t \in \left\{1,\dots,T-1\right\}} (A_{i,t} - A_{i,T}),
\end{align}
where $T$ is the length of the sequential tasks, $A_{i,T}$ is the classification accuracy on the task $i$ after learning the last task $T$.

As analyzed, predicting the \emph{ground truth} set of prompts for each testing sample is a crucial step in PCL. Therefore, we adopt a unique evaluation metric, \textbf{Prompt Retrieval Accuracy (PRA)}, for PCL, which is formally defined as follows:
\begin{align}
    \text{PRA} &= \frac{1}{T} \sum_{i=1}^{T} R_{i,T},
\end{align}
where $R_{i,T}$ is the accuracy of predicting the set of prompts for each testing sample on task $i$ after learning the last task $T$. Besides, we also use \textbf{Selectable Sets of Prompt (SSP)} to represent the total amount of selectable sets of prompts in the pool $\mathcal{P}$. SSP is not only positively correlated with the number of learnable parameters, but it also effectively reflects how LW2G can significantly reduce the selectable amount in baseline methods, thereby benefiting PRA. 

Additionally, we employed two extra metrics in Table \ref{tab:long_task_table_overheads}: the \textbf{\textit{prompt}} and the \textbf{\textit{base}}. (1) The \textit{prompt} is a common storage overhead in the PCL method, and its size is directly proportional to SSP. To elaborate, here we consider an example from the benchmark: CIFAR(Inc10Task10) and the baseline: DualPrompt. In this case, e-prompt is utilized in the first 5 Multi-Head Self-Attention (MHSA) layers of the pre-trained ViT, with each layer incorporating prompts of length 20, consistent with the pre-trained ViT. Each prompt has a dimension of 768. Moreover, since DualPrompt employs the prefix tuning method, which is one of the widely used prompt tuning techniques, it concatenates the prompts to both the key (K) and value (V) in the MHSA mechanism. Therefore, the \textit{prompt} here constitutes 10 $\times$ 5 $\times$ 20 $\times$ 2 = 2000 tensors of 768 dimensions, with each tensor stored as float32. (2) The \textit{base}, unique to the GPCL method, represents an additional storage overhead introduced by LW2G compared to the baseline. The \textit{base} is used to characterize the feature space of old tasks, denoted as $\mathcal{S}_{old}$, and is employed for gradient modification using $\mathcal{S}_{old}$.

\subsection{Main Results}
From Table \ref{tab:main_table1}(\textcolor{orange}{a}), we observe that single-set methods face a performance bottleneck, especially on IMR where task disparity is large. This indicates that the prompt/LoRA set should dynamically grow or not according to task disparities.
From Table \ref{tab:main_table1}(\textcolor{orange} {b,c}), as a plug-in method, LW2G shows consistent improvement across up to 9 existing baselines (including prompt-based and LoRA-based). Specifically, for IMR(Inc20Task10), LW2G outperforms DualPrompt, S-Prompt++, HiDe-Prompt, and CPrompt by 1.38\%, 2.15\%, 0.77\%, and 2.02\% on FAA, respectively; for CIFAR(Inc10Task10), LW2G surpasses C-LoRA, InfLoRA-b5, HiDe-LoRA, DualPrompt, and HiDe-Prompt by 1.72\%, 0.16\%, 1.68\%, 0.92\%, and 1.83\% on FAA, respectively. Additionally, it appears that LW2G brings a significant reducation in anti-forgetting, especially when compared with S-Prompt++, HiDe-Prompt, and CPrompt, showing -0.64\%, -1.91\%, and -1.28\% on FFM, respectively. Furthermore, LW2G also leads to a significant improvement in PRA, with the highest improvement reaching up to 33.39\%, along with a substantial reduction in SSP. For example, DualPrompt w/LW2G only requires 2 sets of prompts compared to the original 10 sets (DualPrompt). From Table \ref{tab:main_table1}(\textcolor{orange}{d}), we report the overall performance gains obtained by employing LW2G as a plug-in method over 9 baselines.

From Table \ref{tab:main_table2}, LW2G achieves improvements over the other two plug-in methods, LGCL and PGP. For instance, compared with PGP, LW2G brings about 1.38\% and 13.00\% improvement on FAA and PRA, respectively. Moreover, in terms of the number of parameters (simply denoted as SSP), it is reduced from 30 sets to 9 sets. It is noteworthy that LGCL introduces an additional language model during training, posing an unfair comparison with other methods that solely use visual models. However, even in comparison with LGCL, LW2G is still able to achieve consistent improvements.

\subsection{Overhead Analysis}
Learning in long sequential tasks is particularly challenging in continual learning (CL), and as previously analyzed, existing PCL methods add a new prompt/LoRA set for each new task, leading to a growing pool that may cause potential parameter redundancy. Therefore, we discuss the overhead of Baseline and Baseline w/LW2G in two extreme settings: OMNI(Inc10Task30) and OMNI(Inc5Task60). The total overhead is divided into two parts: (i) the learnable prompts, i.e., the \textit{prompt} in Table \ref{tab:long_task_table_overheads}, which has a linear relationship with SSP in Table \ref{tab:main_table1} and Table \ref{tab:main_table2}; (ii) the orthogonal bases, i.e., the \textit{base} in Table \ref{tab:long_task_table_overheads}, which is unique to LW2G and used for the construction of the old feature space and gradient modification, stored as a tensor batch of length 768 in float32 format. Despite the additional orthogonal bases, LW2G dynamically decides whether to grow the prompt set, i.e., learnable parameters, thereby significantly reducing the \textit{overall overhead}. Additionally, we provide \textit{FLOPs} and \textit{running time} to measure the calculation burden. Due to the need for dynamic decisions on whether to grow or not, moderate increases in \textit{FLOPs} and \textit{running time} are introduced. In summary, LW2G achieves performance improvements over the baseline while reducing the number of learnable parameters, with the highest improvement reaching up to 1.76\% on FAA, and only incurring slight additional calculation burden.

\begin{table}[!t]
  \centering
  \caption{Ablation study about the influence of different modules in LW2G. Here we denote variants in LW2G, e.g., ``DGA'' refers to the use of Dynamic Growing Approach within the DualPrompt.}
  \label{tab:ablation_table}
  \resizebox{\columnwidth}{!}{
  \begin{tabular}{llp{0.cm}llll}
    \toprule
    \multicolumn{2}{c}{Modules} && \multicolumn{4}{c}{\textit{IMR(Inc20Task10)}} \\ 
    \cmidrule{1-2} \cmidrule{4-7}
    DGA & CPK && FAA ($\uparrow$) & PRA ($\uparrow$) & FFM ($\downarrow$) & SSP ($\downarrow$) \\
    \midrule
    \xmark  & \xmark  && 67.30 & 45.51 & 6.41 & 10 \\
    \rowcolor{LightCyan}\checkmark & \checkmark && \textbf{68.68} & \textbf{78.90} & \textbf{5.72} & \textbf{2} \\
    \checkmark & \xmark  && 68.35 & 78.90 & 5.92 & 2 \\
    \xmark  & \checkmark && 67.61 & 45.51 & 6.21 & 10 \\
    \bottomrule
  \end{tabular}
  }
\end{table}


\begin{table}[!t]
  \centering
\caption{Different Desgins of DGA with FAA ($\uparrow$) metric on \textit{CIFAR(Inc10Task10)} and \textit{IMR(Inc20Task10)} .} 
\label{tab:variants_on_dsa}
  \resizebox{\columnwidth}{!}{
  \begin{tabular}{lllp{0.cm}ll}
    \toprule
    Methods & CPK & DGA Desgin && \textit{CIFAR} & \textit{IMR} \\
    \midrule
    DualPrompt & \xmark & \xmark && 85.94 & 67.30 \\
    \multirow{4}{*}{\quad w/ LW2G} & \cmark & RND && 85.99 & 66.52\\
    & \cmark & ONE && 84.78 & 65.63\\
    & \cmark & MAX && 86.08 & 66.71\\
    \rowcolor{LightCyan}& \cmark & MIN && \textbf{86.86} & \textbf{68.68}\\
    \midrule
    S-Prompt++ & \xmark & \xmark && 89.25 & 66.03 \\
    \multirow{4}{*}{\quad w/ LW2G} & \cmark & RND && 88.32 & 65.39\\
    & \cmark & ONE && 85.17 & 64.10\\
    & \cmark & MAX && 86.73 & 65.94\\
    \rowcolor{LightCyan}& \cmark & MIN && \textbf{89.32} & \textbf{68.18}\\
    \bottomrule
  \end{tabular}
  }
\end{table}

\begin{figure*}[!t]
  \centering

  \subfloat[Baseline: DualPrompt, Benchmark: \textit{CIFAR(Inc10Task10)}, $\phi=0.5$]{
    \includegraphics[width=0.99\linewidth]{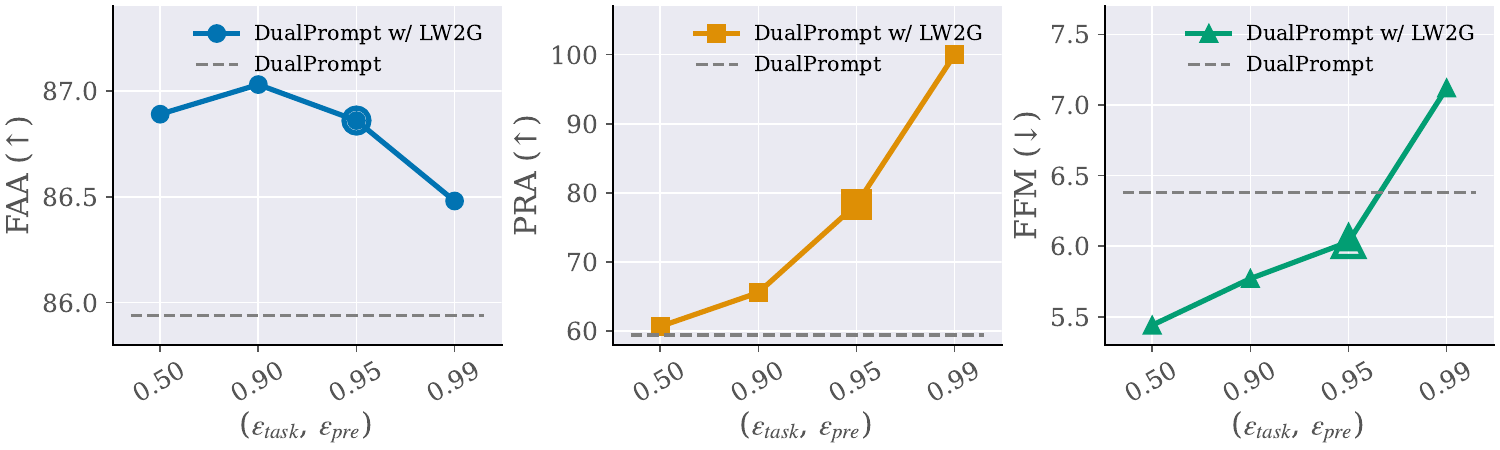}
  }

  \par\vspace{-0.1cm}



  \subfloat[Baseline: DualPrompt, Benchmark: \textit{CIFAR(Inc10Task10)}, $\epsilon_{task}=\epsilon_{pre}=0.95$]{
    \includegraphics[width=0.99\linewidth]{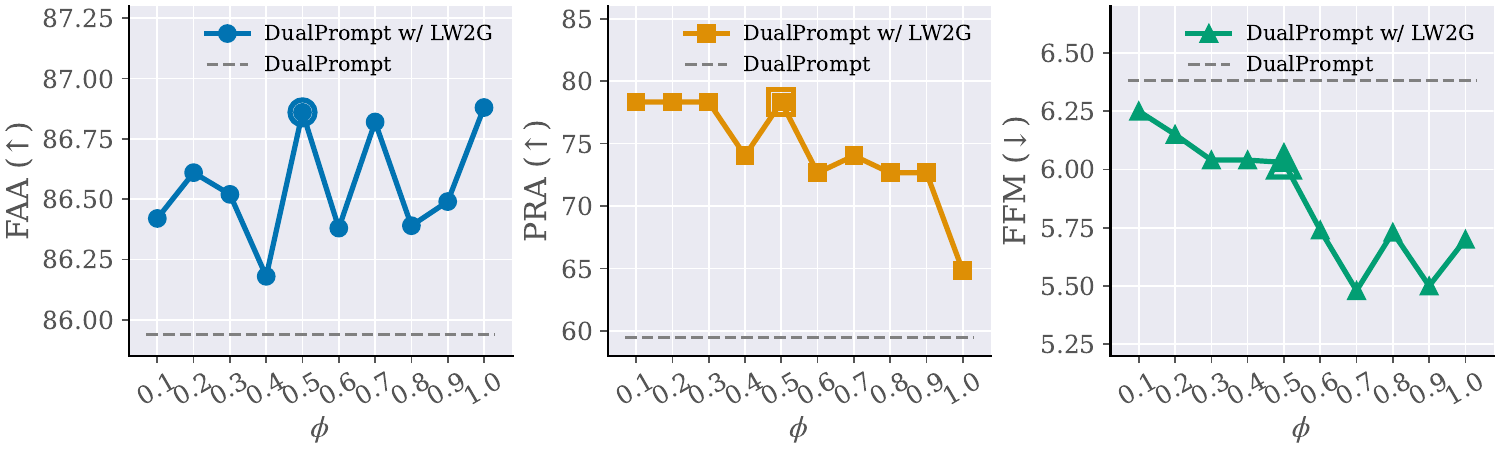}
  }


\vspace{-3mm}
  \caption{
    (a) The impact on FAA, PRA, and FFM when varying $\epsilon_{task}$ and $\epsilon_{pre}$ with a fixed $\phi = 0.5$.
    (b) The impact on FAA, PRA, and FFM when varying $\phi$ with fixed $\epsilon_{task} = \epsilon_{pre} = 0.95$.
  }
  \label{fig:fourplots}
\end{figure*}

\begin{table*}[t!]
  \centering
    \caption{Results under different pre-trained models (PTM), e.g., IBOT-21K, IBOT-1K and DINO-1K, when comparing LW2G with three baselines. The best results are highlighted in bold.}
    \label{tab:appendix_table_ibot21k}
 \vspace{-0.32cm}
  \resizebox{\textwidth}{!}{
  \begin{tabular}
  {p{2.5cm}p{2.0cm}p{1.5cm}p{1.5cm}p{1.3cm}p{1.3cm}p{0.cm}p{1.5cm}p{1.5cm}p{1.3cm}p{1.3cm}}
  \toprule
   \multirow{3}{*}{Method} & \multirow{3}{*}{PTM} & \multicolumn{4}{c}{\emph{CIFAR(Inc10Task10)}} && \multicolumn{4}{c}{\emph{IMR(Inc20Task10)}} \\
  \cmidrule{3-6} \cmidrule{8-11}
   & & FAA ($\uparrow$) & PRA ($\uparrow$) & FFM ($\downarrow$) & SSP ($\downarrow$) && FAA ($\uparrow$) & PRA ($\uparrow$) & FFM ($\downarrow$) & SSP ($\downarrow$) \\
    \midrule
    DualPrompt & \multirow{1}{*}{IBOT-21K} & 74.03 & 72.16 & 15.93 & 10 && 47.96 & 38.62 & 5.36 & 10 \\
    \rowcolor{LightCyan}\quad w/ LW2G && \textbf{74.76} & \textbf{78.33} & \textbf{13.92} & \textbf{3} && \textbf{49.13} & \textbf{64.05} & \textbf{5.33} & \textbf{3} \\
     && \gr{+0.73} & \gr{+6.17} & \bluegr{-2.01} & \bluegr{-7} && \gr{+1.17} & \gr{+25.43} & \bluegr{-0.03} & \bluegr{-7} \\

    S-Prompt++ & \multirow{1}{*}{IBOT-21K} & 78.37 & 75.20 & 9.00 & 10 && 46.20 & 37.77 & 7.01 & 10 \\
    \rowcolor{LightCyan}\quad w/ LW2G && \textbf{78.83} & \textbf{80.31} & \textbf{8.69} & \textbf{3} && \textbf{48.97} & \textbf{71.04} & \textbf{6.30} & \textbf{3} \\
     && \gr{+0.46} & \gr{+5.11} & \bluegr{-0.31} & \bluegr{-7} && \gr{+2.77} & \gr{+33.27} & \bluegr{-0.71} & \bluegr{-7} \\

    HiDe-Prompt & \multirow{1}{*}{IBOT-21K} & 86.12 & 85.02 & 5.98 & 10 && 62.00 & 67.28 & 5.63 & 10 \\
    \rowcolor{LightCyan}\quad w/ LW2G && \textbf{86.40} & \textbf{92.06} & \textbf{5.84} & \textbf{2} && \textbf{63.67} & \textbf{82.18} & \textbf{5.80} & \textbf{3} \\
     && \gr{+0.28} & \gr{+7.04} & \bluegr{-0.14} & \bluegr{-8} && \gr{+1.67} & \gr{+14.90} & \bluegr{-0.17} & \bluegr{-7} \\
    \midrule
    DualPrompt & \multirow{1}{*}{IBOT-1K} & 71.58 & 84.72 & 19.41 & 10 && 56.68 & 38.15 & 5.18 & 10 \\
    \rowcolor{LightCyan}\quad w/ LW2G && \textbf{71.79} & \textbf{84.90} & \textbf{18.99} & \textbf{3} && \textbf{56.89} & \textbf{57.57} & \textbf{5.04} & \textbf{3} \\
     && \gr{+0.21} & \gr{+0.18} & \bluegr{-0.42} & \bluegr{-7} && \gr{+0.21} & \gr{+19.42} & \bluegr{-0.14} & \bluegr{-7} \\

    S-Prompt++ & \multirow{1}{*}{IBOT-1K} & 75.70 & 83.76 & 9.46 & 10 && 52.38 & 39.78 & 7.18 & 10 \\
    \rowcolor{LightCyan}\quad w/ LW2G && \textbf{76.01} & \textbf{84.37} & \textbf{8.91} & \textbf{3} && \textbf{55.82} & \textbf{55.90} & \textbf{7.13} & \textbf{3} \\
     && \gr{+0.31} & \gr{+0.61} & \bluegr{-0.55} & \bluegr{-7} && \gr{+3.44} & \gr{+16.12} & \bluegr{-0.05} & \bluegr{-7} \\

    HiDe-Prompt & \multirow{1}{*}{IBOT-1K} & 84.83 & 83.50 & 6.48 & 10 && 64.77 & 67.94 & 6.90 & 10 \\
    \rowcolor{LightCyan}\quad w/ LW2G && \textbf{85.54} & \textbf{88.02} & \textbf{5.75} & \textbf{3} && \textbf{65.15} & \textbf{78.27} & \textbf{4.86} & \textbf{3} \\
     && \gr{+0.71} & \gr{+4.52} & \bluegr{-0.73} & \bluegr{-7} && \gr{+0.38} & \gr{+10.33} & \bluegr{-2.04} & \bluegr{-7} \\
    \midrule
    DualPrompt & \multirow{1}{*}{DINO-1K} & 69.46 & 88.80 & 18.96 & 10 && 52.41 & 38.74 & 5.93 & 10 \\
    \rowcolor{LightCyan}\quad w/ LW2G && \textbf{70.13} & \textbf{89.01} & \textbf{18.03} & \textbf{3} && \textbf{54.22} & \textbf{75.75} & \textbf{5.77} & \textbf{2} \\
     && \gr{+0.67} & \gr{+0.21} & \bluegr{-0.93} & \bluegr{-7} && \gr{+1.81} & \gr{+37.01} & \bluegr{-0.16} & \bluegr{-8} \\

    S-Prompt++ & \multirow{1}{*}{DINO-1K} & 71.36 & 87.60 & 12.38 & 10 && 60.00 & 37.72 & 6.75 & 10 \\
    \rowcolor{LightCyan}\quad w/ LW2G && \textbf{74.62} & \textbf{89.30} & \textbf{10.71} & \textbf{2} && \textbf{65.44} & \textbf{79.35} & \textbf{6.01} & \textbf{5} \\
     && \gr{+3.26} & \gr{+1.70} & \bluegr{-1.67} & \bluegr{-8} && \gr{+5.44} & \gr{+41.63} & \bluegr{-0.74} & \bluegr{-5} \\

    HiDe-Prompt & \multirow{1}{*}{DINO-1K} & 82.89 & 82.05 & 7.45 & 10 && 62.42 & 62.07 & 8.89 & 10 \\
    \rowcolor{LightCyan}\quad w/ LW2G && \textbf{83.58} & \textbf{88.57} & \textbf{7.08} & \textbf{3} && \textbf{64.04} & \textbf{86.43} & \textbf{4.82} & \textbf{2} \\
     && \gr{+0.69} & \gr{+6.52} & \bluegr{-0.37} & \bluegr{-7} && \gr{+1.62} & \gr{+24.36} & \bluegr{-4.07} & \bluegr{-8} \\

  \bottomrule
\end{tabular}
}

\end{table*}

\vspace{-4mm}
\subsection{Ablation Study and Analysis}
\label{Ablation_Study}
\vspace{-2mm}
\subsubsection{The Influence of Each Modules in LW2G}

To verify the effectiveness of different components in our proposed LW2G, ablation experiments are conducted on DualPrompt as baseline and reported in Table \ref{tab:ablation_table}. Overall, optimizing each component yields clear benefits, with all contributing to the robust gains of LW2G. Besides, it is noteworthy that CPK does not reduce SSP, hence the performance improvement solely stems from the enhanced representational capacity of prompts. DGA not only integrates knowledge from multiple tasks into a single set of prompts, thereby enhancing the representational capacity, but importantly, the notable improvement in PRA is attributed to the reduction in the total number of available sets during prompt retrieval, thereby aiding PCL performance.

\vspace{-2mm}
\subsubsection{The Influence of Different Designs in DGA}
While chosing \emph{not to grow}, DGA utilized in LW2G selects the set $(\bm{p}_{*},\bm{k}_{*})$ with the Min-$Z$ from $\mathcal{P}$ when learning task $i$, and learns new knowledge based on this set, adjusting gradient to prevent forgetting of the old knowledge contained in $(\bm{p}_{*},\bm{k}_{*})$. After learning, $(\bm{p}_{*},\bm{k}_{*})$ encompasses both the new knowledge from task $i$ and the existing old knowledge. Here, we explore the impact of different implementations of DGA. In Table \ref{tab:variants_on_dsa}, DGA-RND denotes randomly selecting an old set of prompts from $\mathcal{P}$; DGA-ONE denotes that $\mathcal{P}$ consists of only a single set, implying continuous learning of new knowledge on this set; DGA-MAX indicates selecting the set from $\mathcal{P}$ with the maximum HFC value; and DGA-MIN is the default setting in LW2G. The results clearly demonstrate the superiority of DGA-MIN over other variants, aligning with Theorem \ref{theorem_1}. 
\vspace{-2mm}
\subsubsection{The Influence of Different Hyperparameters}

\noindent {$\epsilon_\text{task}$, $\epsilon_\text{pre}$:} In Gradient Projection Continual Learning (GPCL), $\epsilon$ is usually used to construct the feature space in the SVD. Previous works set it between 0.9 and 0.99. In LW2G, $\epsilon_{task}$ and $\epsilon_{pre}$ are also used for feature space construction (old knowledge and pre-trained knowledge feature space). Thus, we follow the value in \cite{saha2021gradient, qiao2023prompt, zhao2023rethinking} and set these two parameters with the same value. We performed a grid search for appropriate values under different settings. As shown in Figure \ref{fig:fourplots}(a), LW2G consistently bring performance improvement for any of the aforementioned values.

\noindent {$\phi$:} $\phi$ controls the pre-trained knowledge and the acquisition of new task knowledge. We performed a grid search for $\phi$ and the results are shown in Figure \ref{fig:fourplots}(b).

\vspace{-2mm}
\subsubsection{The Influence of Different PTMs}
\label{appendix:other_backbone}

To demonstrate the efficacy of the proposed method under a variety of pre-training paradigms for ImageNet-21K and ImageNet-1K, we further evaluate our LW2G by extending three distinct PTMs, namely IBOT-1K (Zhou et al., 2021), IBOT-21K (Zhou et al., 2021), and DINO-1K (Caron et al., 2021). The results are shown in Table~\ref{tab:appendix_table_ibot21k}, which demonstrate that our LW2G can consistently and significantly improve performance over all baselines.

\begin{figure}[!h]
    \centering
    \vspace{-4mm}
    \includegraphics[width=1\linewidth]{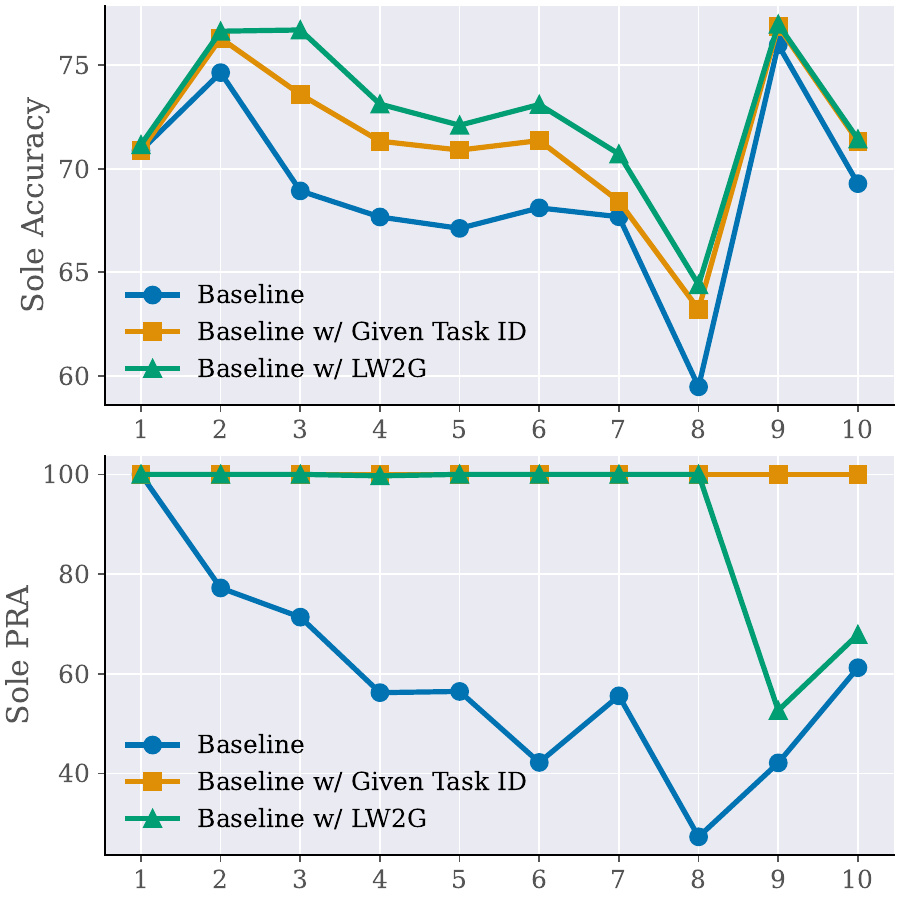}
        \vspace{-4mm}
    \caption{Baseline: DualPrompt. Benchmark: IMR(Inc20Task10). The two figures respectively represent the evaluation results ((a) the accuracy and (b) the PRA) on the $i$-th task immediately after the completion of training for the $i$-th task.}
    \label{label_pic_1}
    \vspace{-6mm}
\end{figure}

\subsection{Intra-Task Knowledge Sharing and Cooperation Via LW2G}
When DGA chooses not to grow, i.e., reuses an old prompt set in the prompt pool, the new task is learned within this old prompt set, thereby injecting new task knowledge into it. At the same time, due to the \textit{orthogonal condition} applied during new task learning, the old knowledge is well preserved. As a result, after learning the new task, the selected old prompt set contains both old and new knowledge, serving as a shared prompt set for both the new and old tasks. This kind of shared strategy not only reduces the total number of prompt sets in the pool, thereby lowering the chance of incorrect predictions caused by selecting a wrong prompt set during inference, but more importantly, it integrates new and old knowledge together, effectively promoting intra-task knowledge sharing and cooperation. This finding is also revealed by \cite{yu2024boosting}, which enhances the inference process by using a MoE mechanism to combine independent knowledge from multiple tasks. Similar to them, our method not only achieves intra-task knowledge sharing, but also reduces parameters through the DGA module.

Specifically, from Figure \ref{label_pic_1}, the y-axis represents the performance solely on task $(i)$ after learning task $(i)$. Both Baseline w/TaskID (where the correct task-wise prompt set is used for each testing sample) and Baseline w/LW2G (which maintains a single prompt set for 8 tasks) achieve 100\% PRA on the first 8 tasks. However, the performance of Baseline w/LW2G exceeds that of Baseline w/TaskID. This suggests that through LW2G, gradually consolidating knowledge from multiple tasks (e.g., task $(i)$ and task $(i+1)$) into a single prompt set is much more effective than using a prompt set that is learned solely on task $(i+1)$. However, it is worth noting that Baseline w/TaskID represents the most ideal scenario in PCL, and the performance of Baseline is significantly lower than Baseline w/TaskID. Therefore, the superiority of Baseline w/LW2G over Baseline stems not only from the improvement in PRA but also from the DGA module consolidating knowledge from multiple tasks into a single prompt, which promotes intra-task knowledge sharing and cooperation.

\begin{table*}[!h]
    \centering
    \caption{Dynamic growing process of DualPrompt w/LW2G on \textit{IMR(Inc20Task10)}.}
    \label{tab:dualprompt_lw2g}
    \renewcommand{\arraystretch}{1.3}
    \setlength{\tabcolsep}{5pt}
    \footnotesize

    \begin{subtable}{\textwidth}
        \centering
        \caption{Task 1--5}
        \begin{tabularx}{\textwidth}{l|*{5}{C}}
            \toprule
            {Task} & {1} & {2} & {3} & {4} & {5} \\
            \midrule
            {HFC} 
            & /
            & \begin{tabular}[c]{@{}c@{}}$\text{HFC}_1 = 13.90$ \\ $\text{HFC}_1^{\text{pre}} = 40.23$\end{tabular}
            & \begin{tabular}[c]{@{}c@{}}$\text{HFC}_1 = 20.22$ \\ $\text{HFC}_1^{\text{pre}} = 40.80$\end{tabular}
            & \begin{tabular}[c]{@{}c@{}}$\text{HFC}_1 = 25.09$ \\ $\text{HFC}_1^{\text{pre}} = 41.50$\end{tabular}
            & \begin{tabular}[c]{@{}c@{}}$\text{HFC}_1 = 29.15$ \\ $\text{HFC}_1^{\text{pre}} = 42.92$\end{tabular} \\
            \midrule
            {Min-Z} 
            & /
            & $Z_1 = -26.33 < 0$
            & $Z_1 = -20.58 < 0$
            & $Z_1 = -16.41 < 0$
            & $Z_1 = -13.77 < 0$ \\
            \midrule
            {Option} 
            & \textbf{Grow} $(\bm{p}_1,\bm{k}_1)$
            & \textbf{Reuse} $(\bm{p}_1,\bm{k}_1)$
            & \textbf{Reuse} $(\bm{p}_1,\bm{k}_1)$
            & \textbf{Reuse} $(\bm{p}_1,\bm{k}_1)$
            & \textbf{Reuse} $(\bm{p}_1,\bm{k}_1)$ \\
            \midrule
            {Pool} 
            & $(\bm{p}_1,\bm{k}_1) \!\rightarrow\!$ Task 1
            & $(\bm{p}_1,\bm{k}_1) \!\rightarrow\!$ Task 1,2
            & $(\bm{p}_1,\bm{k}_1) \!\rightarrow\!$ Task 1,2,3
            & $(\bm{p}_1,\bm{k}_1) \!\rightarrow\!$ Task 1--4
            & $(\bm{p}_1,\bm{k}_1) \!\rightarrow\!$ Task 1--5 \\
            \bottomrule
        \end{tabularx}
    \end{subtable}

    \vspace{1.2em}

    \begin{subtable}{\textwidth}
        \centering
        \caption{Task 6--10}
        \begin{tabularx}{\textwidth}{l|*{5}{C}}
            \toprule
            {Task} & {6} & {7} & {8} & {9} & {10} \\
            \midrule
            {HFC} 
            & \begin{tabular}[c]{@{}c@{}}$\text{HFC}_1 = 32.85$ \\ $\text{HFC}_1^{\text{pre}} = 42.78$\end{tabular}
            & \begin{tabular}[c]{@{}c@{}}$\text{HFC}_1 = 36.35$ \\ $\text{HFC}_1^{\text{pre}} = 41.85$\end{tabular}
            & \begin{tabular}[c]{@{}c@{}}$\text{HFC}_1 = 39.39$ \\ $\text{HFC}_1^{\text{pre}} = 42.42$\end{tabular}
            & \begin{tabular}[c]{@{}c@{}}$\text{HFC}_1 = 42.54$ \\ $\text{HFC}_1^{\text{pre}} = 41.37$\end{tabular}
            & \begin{tabular}[c]{@{}c@{}}$\text{HFC}_1 = 42.54$ \\ $\text{HFC}_1^{\text{pre}} = 40.92$ \\
               $\text{HFC}_2 = 13.81$ \\ $\text{HFC}_2^{\text{pre}} = 41.81$\end{tabular} \\
            \midrule
            {Min-Z} 
            & $Z_1 = -9.33 < 0$
            & $Z_1 = -5.5 < 0$
            & $Z_1 = -3.03 < 0$
            & $Z_1 = 1.17 > 0$
            & $Z_2 = -28.00 < 0$ \\
            \midrule
            {Option} 
            & \textbf{Reuse} $(\bm{p}_1,\bm{k}_1)$
            & \textbf{Reuse} $(\bm{p}_1,\bm{k}_1)$
            & \textbf{Reuse} $(\bm{p}_1,\bm{k}_1)$
            & \textbf{Grow} $(\bm{p}_2,\bm{k}_2)$
            & \textbf{Reuse} $(\bm{p}_2,\bm{k}_2)$ \\
            \midrule
            {Pool} 
            & $(\bm{p}_1,\bm{k}_1) \!\rightarrow\!$ Task 1--6
            & $(\bm{p}_1,\bm{k}_1) \!\rightarrow\!$ Task 1--7
            & $(\bm{p}_1,\bm{k}_1) \!\rightarrow\!$ Task 1--8
            & $(\bm{p}_1,\bm{k}_1) \!\rightarrow\!$ Task 1--8; $(\bm{p}_2,\bm{k}_2) \!\rightarrow\!$ Task 9
            & $(\bm{p}_1,\bm{k}_1) \!\rightarrow\!$ Task 1--8; $(\bm{p}_2,\bm{k}_2) \!\rightarrow\!$ Task 9,10 \\
            \bottomrule
        \end{tabularx}
    \end{subtable}
\end{table*}

\subsection{Visualization of the Dynamic Growing Process}
In LW2G, the DGA module determines whether to grow a new set of prompts or reuse an existing set from the prompt sets pool based on the HFC metric, which can measure the hindrance on learning new tasks while maintaining old knowledge under the \textit{orthogonal condition}. We provide a detailed dynamic process in Table \ref{tab:dualprompt_lw2g}(a,b). Before learning each task (except task $1$), LW2G first calculates the HFC value and subsequently decides whether to perform dynamic expansion based on the minimum Z value using Eq.(\ref{tab:get_z}) and Eq.(\ref{tab:decide}).

\vspace{-4mm}
\section{Conclusion}

In this paper, we propose a plug-in method within existing Prompt-based Continual Learning (PCL), called \textit{Learning Whether To Grow} (LW2G). Specifically, LW2G enables PCL to dynamically learn whether to add a new set of prompts for each task (\emph{to grow}) or to utilize an existing set of prompts (\emph{not to grow}) based on the relationships between tasks. Inspired by Gradient Projection-based Continual Learning (GPCL), we leverage the \emph{orthogonal condition} to construct an effective and efficient prompt set pool. Furthermore, we provide a theoretical analysis of hindrance under the \emph{orthogonal condition} in GPCL. Extensive experiments demonstrate the effectiveness of our method. Future work includes extending LW2G from prompt-based continual learning to more general structural-based continual learning methods, exploring its capability to control network expansion in various parameter-efficient fine-tuning methods.

\noindent \textbf{Limitations:} The limitations are two-folds: Firstly, constructing feature spaces and modifying gradients inevitably introduce additional computational overhead, which may limit its practical deployment. Developing more efficient feature space construction techniques could be explored to further enhance the applicability of LW2G. Secondly, the current design is mainly validated on prompt-based expansion; how to generalize the same idea to more complex structural expansion scenarios (e.g., hybrid architectures) remains an open question and will be investigated in future work.




\section{Data Availability Statement}

Code is available at \url{https://github.com/RAIAN08/LW2G}.

\bibliographystyle{plain}      
\bibliography{paper}   

\end{document}